\documentclass[11pt,a4paper]{article}

\usepackage[utf8]{inputenc}
\usepackage[T1]{fontenc}
\usepackage{amsmath,amssymb}
\usepackage{graphicx}
\usepackage{booktabs}
\usepackage{natbib}
\usepackage{hyperref}
\usepackage{xcolor}
\usepackage{enumitem}
\usepackage{multirow}
\usepackage{longtable}
\usepackage{caption}
\usepackage{float}
\usepackage[margin=1in]{geometry}
\usepackage{microtype}

\hypersetup{
  colorlinks=true,
  linkcolor=blue!60!black,
  citecolor=blue!60!black,
  urlcolor=blue!60!black,
}

\graphicspath{{figures/}}

\newcommand{\numconversations}{4{,}182}
\newcommand{\numexchanges}{14{,}340}
\newcommand{\numdistilledobjects}{12{,}427}
\newcommand{\compressionratio}{11}
\newcommand{\avgverbatimtokens}{371}
\newcommand{\avgdistilledtokens}{38}
\newcommand{\numprojects}{6}
\newcommand{\compressionratiotext}{\compressionratio$\times$}
\newcommand{\compressioncostfraction}{$\frac{1}{\compressionratio{}}$}
\newcommand{\numqueries}{201}
\newcommand{\nummodes}{14}
\newcommand{\numconfigs}{107}
\newcommand{\topk}{10}

\newcommand{\numpuremodes}{5}
\newcommand{\numpureconfigs}{44}
\newcommand{\numcrosslayermodes}{5}
\newcommand{\numcrosslayerconfigs}{74}

\newcommand{\numhybridconfigs}{56}
\newcommand{\numcrosslayerevalconfigs}{63}

\newcommand{\verbatimHnswMRR}{0.645}
\newcommand{\verbatimHnswGrade}{2.002}

\newcommand{\verbatimExactMRR}{0.638}
\newcommand{\verbatimBmokapiMRR}{0.719}
\newcommand{\verbatimBmftsMRR}{0.745}
\newcommand{\bestVerbatimMRR}{0.745}
\newcommand{\bestVerbatimLabel}{Full Text / BM25-FTS}
\newcommand{\bestDistilledMRR}{0.717}
\newcommand{\bestDistilledLabel}{Distill Core+Rooms / Exact / Weighted}
\newcommand{\bestDistilledMRRPct}{96}

\newcommand{\bestDistilledGradePct}{93}
\newcommand{\bestDistilledPAtOnePct}{97}
\newcommand{\bestDistilledNDCGPct}{91}
\newcommand{\bestCrossLayerMRR}{0.759}
\newcommand{\bestCrossLayerLabel}{Cross: BM25(V)+HNSW(D) / BM25-FTS + HNSW / CombMNZ}

\newcommand{\bestCrossLayerMRRPct}{102}
\newcommand{\closestDistilledGrade}{1.992}
\newcommand{\closestDistilledGradeLabel}{Distill Core+Rooms / HNSW / Additive}
\newcommand{\closestDistilledGradePct}{100}
\newcommand{\spreadverbatim}{0.108}

\newcommand{\gradeOnePct}{34}

\newcommand{\gradeThreePct}{45}
\newcommand{\vectorMRRmin}{0.559}

\newcommand{\keywordMRRmax}{0.745}
\newcommand{\qtExactDistillCFHnswWt}{2.08}
\newcommand{\qtExactVerbatimHnsw}{1.93}
\newcommand{\qtConceptualVerbatimHnsw}{2.08}
\newcommand{\qtConceptualDistillCFHnswRRF}{1.96}

\newcommand{\ngraders}{5}
\newcommand{\fleissk}{0.175}
\newcommand{\fleisskinterp}{Slight}
\newcommand{\fleissn}{3{,}672}

\newcommand{\bestpairnameA}{Phi-3.5-Mini}
\newcommand{\bestpairnameB}{Qwen3-8B}
\newcommand{\bestpairkappa}{0.260}
\newcommand{\bestpairinterp}{Fair}
\newcommand{\worstpairnameA}{Mistral-7B}
\newcommand{\worstpairnameB}{Phi-3.5-Mini}
\newcommand{\worstpairkappa}{0.131}
\newcommand{\worstpairinterp}{Slight}

\newcommand{\totalcomparisons}{40}
\newcommand{\totalsig}{20}
\newcommand{\totalnotsig}{20}
\newcommand{\bonferronialpha}{0.00125}
\newcommand{\mindcohen}{0.031}
\newcommand{\maxdcohen}{0.756}
\newcommand{\nmediumeffects}{18}
\newcommand{\hnswsig}{0}
\newcommand{\hnswtotal}{10}
\newcommand{\hnswnotsig}{10}

\newcommand{\exactsig}{0}
\newcommand{\exacttotal}{10}
\newcommand{\exactnotsig}{10}

\newcommand{\bmokapisig}{10}
\newcommand{\bmokapitotal}{10}
\newcommand{\bmokapinotsig}{0}

\newcommand{\bmftssig}{10}
\newcommand{\bmftstotal}{10}
\newcommand{\bmftsnotsig}{0}

\newcommand{\totalGradedPairs}{214{,}519}
\newcommand{\consensusNonUnanimous}{3{,}488}
\newcommand{\consensusNoMajority}{741}
\newcommand{\consensusWeakMajority}{1{,}739}
\newcommand{\consensusStrongMajority}{1008}
\newcommand{\consensusRemainingPairs}{3{,}113}
\newcommand{\consensusUnanimous}{366}
\newcommand{\disagreementPct}{90.5}
\newcommand{\parseFailurePct}{0.0}
\newcommand{\humanGradesTotal}{108}
\newcommand{\pilotGrades}{23}
\newcommand{\poolGrades}{85}
\newcommand{\poolAgreementNum}{82}
\newcommand{\poolAgreementDenom}{85}
\newcommand{\poolAgreementPct}{96.5}
\newcommand{\maxGradeDisagreement}{1}
\newcommand{\acceptanceAgreementPct}{85}
\newcommand{\acceptanceKappaMin}{0.61}
\newcommand{\sampleSizeReq}{50}
\newcommand{\pilotSize}{100}
\newcommand{\pilotInitialNum}{17}
\newcommand{\pilotInitialDenom}{23}
\newcommand{\pilotInitialPct}{74}
\newcommand{\pilotCalibratedNum}{22}
\newcommand{\pilotCalibratedDenom}{23}
\newcommand{\pilotCalibratedPct}{95.7}
\newcommand{\opusModel}{claude-opus-4-6}
\newcommand{\opusHigherPct}{67.8}
\newcommand{\opusSamePct}{17.6}
\newcommand{\opusLowerPct}{14.6}
\newcommand{\graderParamRange}{7--12B}
\newcommand{\graderTemperature}{0.0}
\newcommand{\graderQuantization}{Q4\_K\_M}
\newcommand{\graderFormat}{GGUF}
\newcommand{\verbatimTruncChars}{1{,}500}
\newcommand{\distilledAvgChars}{200}
\newcommand{\queryFilterMinChars}{20}
\newcommand{\queryFilterMaxChars}{1{,}000}
\newcommand{\oversamplingFactor}{2}
\newcommand{\queryRandomSeed}{42}
\newcommand{\exchangeMinChars}{100}
\newcommand{\exchangeMaxPlies}{20}
\newcommand{\messageTruncChars}{4{,}000}

\newcommand{\bestVSolves}{122}
\newcommand{\bestDSolves}{119}
\newcommand{\bestOverlap}{77}
\newcommand{\bestVOnly}{45}
\newcommand{\bestDOnly}{42}

\newcommand{\oracleVOnly}{4}
\newcommand{\oracleDOnly}{32}

\newcommand{\oracleNeither}{10}
\newcommand{\oracleUnion}{191}
\newcommand{\oracleUnionPct}{95.0}
\newcommand{\oracleVConfigs}{4}
\newcommand{\oracleDConfigs}{40}
\newcommand{\oracleDOnlyConceptual}{15}
\newcommand{\oracleDOnlyPhrase}{11}
\newcommand{\oracleDOnlyExact}{6}

\newcommand{\vocabSurvivalK}{15}
\newcommand{\vocabSurvivalRate}{27.0\%}

\newcommand{\queryOverlapRetention}{96.8\%}

\newcommand{\coreFieldMeanIDF}{3.13}
\newcommand{\contextFieldMeanIDF}{5.74}
\newcommand{\contextIDFRatio}{1.84}

\newcommand{\rerankVerbatimBaseline}{122}
\newcommand{\rerankDistillOriginal}{100}
\newcommand{\rerankDistillReranked}{120}
\newcommand{\rerankTotalQueries}{201}
\newcommand{\rerankGained}{36}
\newcommand{\rerankLost}{16}
\newcommand{\rerankGapBefore}{22}
\newcommand{\rerankGapAfter}{2}
\newcommand{\rerankBestGate}{105}
\newcommand{\rerankAlpha}{0.3}
\newcommand{\rerankBeta}{0.7}
\newcommand{\rerankTargetConfig}{Distill Core+Rooms / Exact / Additive}

\newcommand{\bestCaseOneQueryType}{exact term}
\newcommand{\bestCaseOneDiff}{+1.60}
\newcommand{\bestCaseOneDistilledMean}{2.80}
\newcommand{\bestCaseOneVerbatimMean}{1.20}
\newcommand{\bestCaseOneModeLabel}{Distill Core}
\newcommand{\bestCaseOneMechanismLabel}{Exact}
\newcommand{\bestCaseOneFusionLabel}{---}
\newcommand{\bestCaseTwoQueryType}{conceptual}
\newcommand{\bestCaseTwoDiff}{+1.50}
\newcommand{\bestCaseTwoDistilledMean}{2.90}
\newcommand{\bestCaseTwoVerbatimMean}{1.40}
\newcommand{\bestCaseTwoModeLabel}{Distill Core+Files}
\newcommand{\bestCaseTwoMechanismLabel}{HNSW}
\newcommand{\bestCaseTwoFusionLabel}{Additive}
\newcommand{\bestCaseThreeQueryType}{exact term}
\newcommand{\bestCaseThreeDiff}{+1.50}
\newcommand{\bestCaseThreeDistilledMean}{1.60}
\newcommand{\bestCaseThreeVerbatimMean}{0.10}
\newcommand{\bestCaseThreeModeLabel}{Distill Core+Files}
\newcommand{\bestCaseThreeMechanismLabel}{Exact}
\newcommand{\bestCaseThreeFusionLabel}{Additive}
\newcommand{\bestCaseFourQueryType}{conceptual}
\newcommand{\bestCaseFourDiff}{+1.30}
\newcommand{\bestCaseFourDistilledMean}{2.60}
\newcommand{\bestCaseFourVerbatimMean}{1.30}
\newcommand{\bestCaseFourModeLabel}{Distill Core}
\newcommand{\bestCaseFourMechanismLabel}{HNSW}
\newcommand{\bestCaseFourFusionLabel}{---}
\newcommand{\bestCaseFiveQueryType}{phrase}
\newcommand{\bestCaseFiveDiff}{+1.30}
\newcommand{\bestCaseFiveDistilledMean}{2.40}
\newcommand{\bestCaseFiveVerbatimMean}{1.10}
\newcommand{\bestCaseFiveModeLabel}{Distill Core+Rooms}
\newcommand{\bestCaseFiveMechanismLabel}{HNSW}
\newcommand{\bestCaseFiveFusionLabel}{Weighted}
\newcommand{\failureCaseOneQueryType}{phrase}
\newcommand{\failureCaseOneMean}{0.86}
\newcommand{\failureCaseTwoQueryType}{phrase}
\newcommand{\failureCaseTwoMean}{0.90}
\newcommand{\failureCaseThreeQueryType}{phrase}
\newcommand{\failureCaseThreeMean}{0.95}
\newcommand{\failureCaseFourQueryType}{phrase}
\newcommand{\failureCaseFourMean}{0.97}
\newcommand{\failureCaseFiveQueryType}{conceptual}
\newcommand{\failureCaseFiveMean}{0.98}

\title{Structured Distillation for Personalized Agent Memory:\\\compressionratiotext{} Token Reduction with Retrieval Preservation}

\author{
  Sydney Lewis
}

\date{March 2026}

\begin{document}

\maketitle

% ============================================================================
% ABSTRACT
% ============================================================================
\begin{abstract}
Long conversations with an AI agent create a simple problem for one user. The history is useful. Carrying it verbatim is expensive.

We study personalized agent memory: one developer's conversation history with an agent, distilled into a compact retrieval layer for later search. Each exchange is compressed into a compound object with four fields: \texttt{exchange\_core}, \texttt{specific\_context}, thematic \texttt{room\_assignments}, and regex-extracted \texttt{files\_touched}. The searchable distilled text (\texttt{exchange\_core} + \texttt{specific\_context}) averages \avgdistilledtokens{} tokens per exchange. Applied to \numconversations{} conversations (\numexchanges{} exchanges) from \numprojects{} software engineering projects in this single-user corpus, the method reduces average exchange length from \avgverbatimtokens{} to \avgdistilledtokens{} tokens, yielding \compressionratiotext{} compression.

The real test is whether personalized recall survives that compression. Using \numqueries{} recall-oriented queries, \numconfigs{} configurations spanning \numpuremodes{} pure and \numcrosslayermodes{} cross-layer search modes, and \ngraders{} large language model graders (\totalGradedPairs{} consensus-graded query-result pairs), we compare search over distilled and verbatim text on the same user's historical corpus. The best pure distilled configuration reaches \bestDistilledMRRPct{}\% of the best verbatim MRR (\bestDistilledMRR{} vs \bestVerbatimMRR{}).

The result is mechanism-dependent. All \totalnotsig{} vector search configurations (HNSW, Exact) remain non-significant after Bonferroni correction, while all \totalsig{} BM25 configurations degrade significantly (effect sizes $|d|$=\mindcohen{}--\maxdcohen{}). The best cross-layer retrieval setup, combining verbatim keyword search with distilled vector search, slightly exceeds the best pure verbatim baseline (MRR \bestCrossLayerMRR{}). The signals are complementary.

Structured distillation compresses single-user agent memory without uniformly sacrificing retrieval quality. At \compressioncostfraction{} the context cost, thousands of exchanges fit within a single prompt while the verbatim source remains available for drill-down. We release the implementation and analysis pipeline as open-source software.
\end{abstract}

% ============================================================================
% 1. INTRODUCTION
% ============================================================================
\section{Introduction}

The conversation between a human and an AI agent is asymmetric in a specific way. The human remembers fragments: spatial, associative, recognition-based. The agent remembers nothing.
Every session starts from zero.
The human may recall that ``we fixed that connection pool issue three weeks ago,'' but the agent has no access to that history unless it is explicitly provided.

This paper studies that problem in the single-user setting. One developer accumulates a long history of conversations with an agent across ongoing projects. The question is not whether a generic compression method works for everyone. The question is whether that user's own history can be distilled into a compact memory layer that remains useful later.

Retrieval-augmented generation \citep[RAG;][]{lewis2020rag} approaches address this by loading prior conversation text into context at query time.
A typical conversation exchange in our corpus contains \avgverbatimtokens{} tokens on average.
Loading 500 past exchanges would cost nearly 200{,}000 tokens. That is a significant portion of most context windows, and expensive even where the window allows it.

A large fraction of those tokens is irrelevant to retrieval: tool output, boilerplate responses, tangential discussion, repeated file contents.
What the human actually remembers is much smaller: the decision made, the parameter changed, the error that got resolved. In our corpus that usually fits in roughly \avgdistilledtokens{} tokens ({\raise.17ex\hbox{$\scriptstyle\sim$}}25 words).

The prevalent response to this cost is \emph{compaction}: when the context window fills, an LLM summarizes the conversation history into a shorter form and discards the original.
This is lossy summarization applied under pressure.
The summarizer has no schema, no extraction targets, and no contract about what survives. It produces whatever a generic ``summarize this'' instruction yields.
In long conversations, compaction compounds: summaries of summaries degrade iteratively, and information from early exchanges is progressively destroyed.
Yet the verbatim conversation is already persisted on the user's machine.
The context that compaction discards is not gone. It is still on disk, unindexed and unused.
What is missing is not better summarization but structured extraction: a method that produces indexed, retrievable working memory from the conversation while the ground truth remains locally available.

The question is simple. Can that missing layer be built for one user's persistent agent history without wrecking retrieval quality?
We use a structured distillation method that turns each exchange into a compound object preserving the retrievable part of the interaction.
We test information preservation through search quality. If the same user's recall queries retrieve results of comparable quality from distilled text and verbatim text, the distillation kept what mattered.

Our contributions are:
\begin{enumerate}[noitemsep]
  \item A distillation method for single-user conversational memory, producing structured compound objects from conversation exchanges and achieving \compressionratiotext{} compression.
  \item A \numconfigs{}-configuration evaluation of personalized recall across \numpuremodes{} pure and \numcrosslayermodes{} cross-layer search modes. The best pure distilled configuration preserves \bestDistilledMRRPct{}\% of verbatim MRR. Vector search holds up. BM25 does not. Cross-layer fusion slightly exceeds the best pure verbatim baseline.
  \item A concrete memory budget result: 1{,}000 conversation exchanges fit in $\sim$39{,}000 tokens instead of $\sim$407{,}000.
  \item A two-tier architecture for personalized agent memory that separates index from display. Distilled text routes the query. The user still reads the original verbatim exchange.
\end{enumerate}

% ============================================================================
% 2. RELATED WORK
% ============================================================================
\section{Related Work}

This section is selective, not exhaustive. The point is to locate the closest neighboring lines of work, not to survey every memory or summarization system.

\subsection{Agent Memory Systems}

Persistent memory for LLM agents is now a systems problem, not a thought experiment. MemGPT \citep{packer2023memgpt} frames memory management as a hierarchy in which older context is pushed out of the immediate window and later retrieved. Our setting is narrower and more concrete. We study one user's accumulated conversation history with an agent and ask whether that history can be distilled into a compact retrieval layer without breaking later recall.

The representation also differs. We distill each resolved exchange into a compound object with a summary field, a distinguishing technical detail, thematic room assignments, and file references, all tied back to the original verbatim source. The distilled text is an index, not the user-facing artifact. Search returns the original conversation. That separation between retrieval representation and displayed evidence is central here.

\subsection{Conversation Summarization}

Dialogue summarization is well-studied \citep{feng2022survey}. Extractive methods select verbatim segments. Abstractive methods generate novel text. Running-summary approaches maintain a rolling digest of older messages.

This paper asks a narrower question. If each conversation exchange is distilled into a structured object and the verbatim source is kept for drill-down, how much retrieval quality survives the compression? We are not trying to produce a polished running summary. We paraphrase the decision, but we keep the technical handles verbatim. The surviving vocabulary principle constrains the LLM to reuse the participants' phrasing rather than paraphrase freely. The exchange core keeps the terms the participants settled on. The specific context keeps exact values such as error messages, parameter names, and file paths.

\subsection{Context Compression}

Prompt compression techniques such as LLMLingua \citep{jiang2023llmlingua} reduce token counts at inference time by removing tokens with low information content. They operate at runtime and optimize a single inference call.

This is a different problem. We distill offline into a searchable object. The output is structured, not merely shorter text, and the distilled text is retrievable in its own right through both keyword and semantic search.

This paper is narrower than the generic summarization literature in another way. The target is not a public benchmark or a cross-user corpus. It is one user's accumulated conversation history with an agent, treated as a personalized memory system. The relevant question is whether that compressed representation stays searchable enough later.

% ============================================================================
% 3. METHOD
% ============================================================================
\section{Method}

\subsection{Exchange Segmentation}

We adopt a ply model for conversation structure.
A user message constitutes one half-ply; the agent's response completes the ply, forming one exchange.
A new exchange begins when a user message follows a non-empty assistant response with substantive text beyond tool invocations.
This structural boundary usually coincides with task completion. Claude Code conversations follow request-response cycles where users issue new requests after receiving results, so exchange boundaries align with natural topic transitions without requiring content-based detection.

Empty tool-use round-trips (where the assistant invokes a tool and the user provides no substantive text) do not start a new exchange; the current exchange remains open until the assistant produces a substantive response.
Exchanges shorter than \exchangeMinChars{} characters are filtered as trivial; exchanges exceeding \exchangeMaxPlies{} plies are split at fixed intervals.

\subsection{Structured Distillation}

Each exchange is distilled into a compound object with four components (Table~\ref{tab:components}).
The first two (\texttt{exchange\_core}, \texttt{specific\_context}) are LLM-generated; \texttt{room\_assignments} are LLM-generated thematic placements; \texttt{files\_touched} is regex-extracted from the raw exchange text, not generated by the LLM:

\begin{table}[htbp]
\centering
\caption{Components of a distilled object.}
\label{tab:components}
\small
\begin{tabular}{llp{4.2cm}l}
\toprule
Component & Source & Content & Analogy \\
\midrule
\texttt{exchange\_core} & LLM & What was accomplished (1--2 sentences) & Commit message \\
\texttt{specific\_context} & LLM & One distinguishing technical detail & The diff \\
\texttt{room\_assignments} & LLM & 1--3 thematic rooms (type, key, label) & Directory tree \\
\texttt{files\_touched} & Regex & File paths referenced in exchange & Changed files \\
\bottomrule
\end{tabular}
\end{table}

Each object also carries a back-reference to its source: \texttt{conversation\_id}, \texttt{ply\_start}, and \texttt{ply\_end}, enabling drill-down to the original verbatim text.
The distilled text is never shown to the user.
It serves as a retrieval routing artifact: all search modes return the original verbatim conversation text.
Distillation determines which conversations surface and in what rank order; the user always reads the unmodified original.

The \emph{surviving vocabulary} principle drives the distillation. The LLM is told to reuse the specific terms in the exchange, not invent synonyms or paraphrase freely.
When participants settle on a term such as ``connection pool timeout'' rather than ``make connections better,'' that term should survive into the distilled object. Precise terminology is what future queries match.
Only \vocabSurvivalRate{} of the top-\vocabSurvivalK{} highest-IDF tokens per verbatim exchange appear in the corresponding distilled text. Most exchange-specific rarities, such as error strings, file paths, and unique identifiers, are lost.
But \queryOverlapRetention{} of query vocabulary is retained in the distilled corpus: the tokens users actually search for survive distillation, while the tokens lost are exchange-specific and rarely queried.
The \texttt{specific\_context} field carries higher-IDF vocabulary than \texttt{exchange\_core} (mean IDF \contextFieldMeanIDF{} vs.\ \coreFieldMeanIDF{}, a \contextIDFRatio{}$\times$ ratio), confirming that it captures the more discriminative terms.

The distilled text is constructed as:
\begin{verbatim}
  distill_text = f"{exchange_core}\n{specific_context}"
\end{verbatim}
This concatenation averages \avgdistilledtokens{} tokens versus \avgverbatimtokens{} for the verbatim exchange. The reported \compressionratiotext{} ratio is computed from corpus totals, not from per-item averages. The two differ because \numdistilledobjects{} distilled objects are produced from \numexchanges{} exchanges via ply splitting. All token counts use the cl100k\_base encoding \citep{tiktoken2023}.

The evaluated corpus was distilled by Claude Haiku 4.5 \citep{anthropic2025haiku45}, invoked via the Claude CLI (\texttt{claude --print --model haiku}), producing all \numdistilledobjects{} palace objects.
The distilled object count (\numdistilledobjects{}) is lower than the exchange count (\numexchanges{}) because the \exchangeMinChars{}-character minimum filter removes more exchanges than ply splitting (at \exchangeMaxPlies{} plies) adds.
The exact batch prompt is provided in Appendix~\ref{app:distillation-prompt}.

\subsection{Embedding and Indexing}

Distilled text is embedded using all-MiniLM-L6-v2 \citep{reimers2019sentencebert,wang2020minilm}, producing 384-dimensional vectors.
We chose this model for deployment accessibility: at 22M parameters it runs on CPU without a GPU, matching the single-developer target environment; larger embedding models would improve semantic matching at the cost of hardware requirements.
These are stored in a FAISS (Facebook AI Similarity Search) \citep{johnson2019faiss} \texttt{IndexFlatL2} index, separate from the verbatim conversation index.
Separate indices are maintained because of a density mismatch: verbatim chunks average $\sim$\verbatimTruncChars{} characters while distilled objects average $\sim$\distilledAvgChars{} characters.
Each distilled object produces a single vector (no chunking is needed given the short text length).

A BM25 index \citep{robertson2009bm25} is built for each object from the concatenation of \texttt{exchange\_core}, \texttt{specific\_context}, \texttt{files\_touched}, and room metadata (\texttt{room\_key}, \texttt{room\_label}).
The BM25 index is constructed lazily in memory at query time.

\subsection{Search Configurations}

We define \nummodes{} search modes organized into three families.
The evaluation covers the \numpuremodes{} pure modes (\numpureconfigs{} configurations) and \numcrosslayermodes{} cross-layer modes (\numcrosslayerevalconfigs{} configurations), for \numconfigs{} configurations in total.
Hybrid modes define the remaining design space and are left to future work.

\paragraph{Pure modes (5 modes, 44 configs).}
A single retrieval mechanism operates on a single text layer.
\begin{enumerate}[noitemsep]
  \item \textbf{Full Text}: Raw verbatim conversation text (baseline).
  \item \textbf{Distill Core}: \texttt{exchange\_core} + \texttt{specific\_context} only.
  \item \textbf{Distill Core+Files}: Core text + \texttt{files\_touched}.
  \item \textbf{Distill Core+Rooms}: Core text + room metadata.
  \item \textbf{Distill All Facets}: Core text + files + rooms.
\end{enumerate}
Each mode uses 4 mechanisms: \textbf{HNSW} (Hierarchical Navigable Small World \citep{malkov2020hnsw}; approximate nearest neighbor via DuckDB's \texttt{vss} extension \citep{duckdb2024vss}), \textbf{Exact} (brute-force cosine similarity), \textbf{BM25-Okapi} ($k_1$=1.5, $b$=0.75), and \textbf{BM25-FTS} (DuckDB full-text search).
Single-field modes use pass-through (identity) fusion; multi-field modes use 3 fusion strategies (Reciprocal Rank Fusion, RRF; \citealt{cormack2009reciprocal}), Weighted, and Additive.

\paragraph{Cross-layer modes (5 modes, 74 configs).}
Two retrieval mechanisms operate on \emph{different} text layers at the same time, one on verbatim text and one on distilled text, and then fuse their ranked lists:
\begin{enumerate}[noitemsep]
  \item \textbf{Cross: BM25(V)+HNSW(D)}: BM25 keyword search on verbatim text fused with HNSW vector search on distilled text.
  \item \textbf{Cross: BM25(V)+HNSW(D) rev}: Same signals, reversed weighting.
  \item \textbf{Cross: HNSW(V)+BM25(D)}: HNSW on verbatim fused with BM25 on distilled.
  \item \textbf{Cross: HNSW(V)+BM25(D) rev}: Reversed weighting.
  \item \textbf{Cross: 3-Signal}: BM25 on both verbatim and distilled, plus HNSW on distilled.
\end{enumerate}
Each uses 2 compound mechanisms (e.g., BM25-Okapi+HNSW, BM25-FTS+HNSW) and 5--9 fusion strategies including CombMNZ \citep{fox1994combination}, Max, and weighted interpolations at signal weights from 50\% to 95\% (denoted W50--W95).

\paragraph{Hybrid modes (4 modes, 56 configs).}
Two retrieval mechanisms operate on the \emph{same} text layer, combining keyword and semantic search:
\begin{enumerate}[noitemsep]
  \item \textbf{Hybrid Raw / rev}: HNSW + BM25 on verbatim text (forward and reversed weighting).
  \item \textbf{Hybrid Core / rev}: HNSW + BM25 on distilled core text (forward and reversed weighting).
\end{enumerate}
Same compound mechanisms and fusion strategies as cross-layer.

The full design space spans \numpureconfigs{} pure + \numcrosslayerconfigs{} cross-layer + \numhybridconfigs{} hybrid configurations.
The \numconfigs{} evaluated configurations span pure and cross-layer families; not all fusion strategies apply to all modes.
Statistical comparisons for pairwise $t$-tests use the \totalcomparisons{} pure-mode comparisons where distilled and verbatim modes share the same mechanism, with Bonferroni correction: $\alpha$ = 0.05/\totalcomparisons{} = \bonferronialpha{}.

For each configuration, the top \topk{} results are retrieved per query.

% ============================================================================
% 4. EXPERIMENTAL SETUP
% ============================================================================
\section{Experimental Setup}

\subsection{Corpus}

% Auto-generated from publication_v2/tables/table5_corpus_stats.tex
% Table 5: Corpus Statistics
\begin{table}[htbp]
\centering
\caption{Corpus statistics from 6 months of production AI coding sessions across 6 projects.
Token counts based on cl100k\_base encoding, measured on corpus totals (overlapping conversations only).}
\label{tab:corpus}
\begin{tabular}{lr}
\toprule
Statistic & Value \\
\midrule
Total conversations & 4,182 \\
Total exchanges & 14,340 \\
Distilled objects & 12,427 \\
Avg.\ verbatim length & 371 tokens \\
Avg.\ distilled length & 38 tokens \\
Compression ratio & 11$\times$$^*$ \\
Projects covered & 6 \\
\bottomrule
\multicolumn{2}{l}{\scriptsize $^*$Computed from corpus totals (14,340 $\times$ 371 / 12,427 $\times$ 38); per-item average ratio is 9.9$\times$.}
\end{tabular}
\end{table}

The corpus (Table~\ref{tab:corpus}) consists of \numconversations{} conversations from one developer's Claude Code sessions across 6 software engineering projects spanning different domains.
The conversations cover ordinary software engineering work: debugging, feature implementation, code review, architecture discussion, and configuration.
This single-user corpus is not a convenience sample standing in for a population benchmark. It is the intended deployment setting: persistent memory over one developer's long-running agent history.

We refer to the structured distilled objects as \emph{palace objects} due to their organization within a spatial memory palace architecture, where each object is assigned to thematic ``rooms.''
Room assignments are LLM-generated during distillation: the model selects 1--3 rooms per exchange from three types (\texttt{file}, \texttt{concept}, \texttt{workflow}), each with a free-form key and label.
Room keys are not constrained to a predefined vocabulary; the LLM produces them based on the exchange content (e.g., \texttt{concept:retry\_timeout}, \texttt{file:auth\_middleware}, \texttt{workflow:ci\_pipeline}).
Deduplication is by deterministic hash of the \texttt{(room\_type, room\_key, project\_id)} triple, so identical keys from different exchanges map to the same room.

\subsection{Query Set}

\numqueries{} evaluation queries were constructed as recall probes over this user's own history, across three types:
\begin{itemize}[noitemsep]
  \item \textbf{Conceptual} (abstract topics): e.g., broad technical concepts
  \item \textbf{Phrase} (multi-word patterns): e.g., multi-word library or technique references
  \item \textbf{Exact term} (specific identifiers): e.g., file names, database identifiers
\end{itemize}

\paragraph{Query construction process.}
Candidate text was drawn via stratified random sampling (seed \queryRandomSeed{}) across the \numprojects{} project groups, weighted by corpus proportion.
Within each group, user and assistant messages were sampled in equal proportions, filtered to \queryFilterMinChars{}--\queryFilterMaxChars{} characters, and stripped of tool output, system tags, markup, interrupted requests, and code blocks.
\oversamplingFactor{}$\times$ oversampling relative to target counts provided surplus candidates.
Candidate queries were generated from these sampled messages, then manually curated for answerability, uniqueness of intent, and clarity. The final evaluation set was stratified to preserve project-group proportions and the target mix of conceptual, phrase, and exact-term queries.

Up to \topk{} results were retrieved per query per configuration.
One query returned no BM25 matches in some configurations, reducing those to 199--200 queries; a small number of queries returned fewer than \topk{} results.
The final dataset spans \numconfigs{} configurations, yielding \totalGradedPairs{} consensus-graded query-result pairs (\numqueries{} queries $\times$ up to \topk{} results $\times$ \numconfigs{} configurations, minus missing results).

\subsection{Grading Methodology}

Five local LLM graders (\graderParamRange{} parameters) independently assessed each query-result pair:
\begin{itemize}[noitemsep]
  \item Qwen3-8B \citep{qwen2025qwen3}
  \item Phi-3.5-Mini (based on \citealt{abdin2024phi3})
  \item Mistral-7B \citep{jiang2023mistral}
  \item Yi-1.5-9B \citep{young2024yi}
  \item InternLM2.5-7B \citep{cai2024internlm2}
\end{itemize}

Graders used a 0--3 relevance scale following TREC (Text Retrieval Conference) conventions \citep{voorhees2005trec}:
\begin{itemize}[noitemsep]
  \item \textbf{0 --- Irrelevant}: The result has nothing to do with the query. Different topic entirely.
  \item \textbf{1 --- Related}: The result is on a related topic but does not answer the query. It may mention similar concepts or share terminology, but a user would not find what they were looking for.
  \item \textbf{2 --- Highly relevant}: The result contains an answer or useful information for the query, but the answer may be unclear, incomplete, or buried among other content.
  \item \textbf{3 --- Perfectly relevant}: The result is dedicated to the query topic and directly provides the information sought.
\end{itemize}

Graders were instructed to judge semantic relevance rather than keyword matching, and to consider what the original user would actually be trying to recall by issuing the query.
The exact grading prompt is provided in Appendix~\ref{app:grading-prompt}.

Consensus was computed via majority voting across the 5 graders; ties were broken conservatively (lower grade).
For the \consensusNoMajority{} pairs where no grader achieved a majority (2/5 vote concentration), a calibrated Claude Opus grader served as a 6th vote, validated against \humanGradesTotal{} human grades (Section~\ref{sec:human-validation}).
The final consensus uses 6 graders for these \consensusNoMajority{} pairs; the remaining \consensusRemainingPairs{} pairs already have a 5-grader majority (\consensusWeakMajority{} weak, \consensusStrongMajority{} strong, and \consensusUnanimous{} unanimous) and retain that consensus.

\parseFailurePct{}\% of individual grader responses failed to parse (malformed JSON, refusals, or empty output).
Parse failures were distributed across all five graders and did not correlate with search mode or result type; the failures are treated as missing at random.
The remaining responses produced consensus grades for all \numqueries{} queries across all \numconfigs{} configurations.

Verbatim snippets were truncated at \verbatimTruncChars{} characters; palace snippets (averaging $\sim$\distilledAvgChars{} characters) were never truncated.
This differential truncation may systematically disadvantage verbatim results.
All five graders ran as \graderQuantization{} \graderFormat{} (GGML Unified Format) quantizations via \texttt{llama-server}; quantization affects model behavior and the specific \graderFormat{} files are not versioned with hashes.
The v2 grading prompt omits rank position (see Section~\ref{sec:human-validation} and Appendix~\ref{app:grading-prompt}).

Local LLMs were chosen over API models for three reasons: reproducibility (weights are fixed), cost (zero marginal cost over \totalGradedPairs{} items), and lack of rate limits.

\subsection{Human Validation via Calibrated Opus Grader}
\label{sec:human-validation}

With \disagreementPct{}\% of pairs showing some disagreement across \ngraders{} graders, manual grading of all disagreements is infeasible.
We adopted a simple consensus triage. Of \consensusNonUnanimous{} non-unanimous pairs, only the \consensusNoMajority{} with no majority need a 6th grader. These are the 2-2-1 splits where a new vote can create a 3-vote majority from nothing.
The \consensusWeakMajority{} weak-majority (3/5) and \consensusStrongMajority{} strong-majority (4/5) pairs are unaffected by a 6th vote.

Claude Opus (\opusModel{}) served as the calibrated 6th grader.
A \pilotSize{}-pair pilot (selected by rank-1 footprint across all \numconfigs{} configurations, spanning both verbatim and distilled modes) was graded by Opus and compared against \pilotInitialDenom{} human grades.
Initial agreement was \pilotInitialNum{}/\pilotInitialDenom{} (\pilotInitialPct{}\%).
The problem was clear. Opus penalized truncated snippets even when the conversation itself was on-topic.
We corrected for that. The grader was told to judge whether the \emph{conversation} addresses the query topic, not whether the visible snippet contains the full answer. Snippets are excerpts from long conversations, so a truncated response or a matching user question can still show that the conversation is relevant.
Post-calibration agreement: \pilotCalibratedNum{}/\pilotCalibratedDenom{} (\pilotCalibratedPct{}\%).

Opus then graded all \consensusNoMajority{} no-majority pairs with the calibrated lens.
Opus graded higher than the 7B consensus on \opusHigherPct{}\% of pairs, the same on \opusSamePct{}\%, and lower on \opusLowerPct{}\%.
Opus's vote resolved all \consensusNoMajority{} ties.

Human validation (\humanGradesTotal{} grades: \pilotGrades{} pilot + \poolGrades{} from the \consensusNoMajority{} pool) confirmed Opus accuracy: \poolAgreementNum{}/\poolAgreementDenom{} agreement on the \consensusNoMajority{} pool (\poolAgreementPct{}\%), with all \the\numexpr\poolAgreementDenom-\poolAgreementNum\relax{} disagreements within $\pm$\maxGradeDisagreement{} grade.
The acceptance criteria (agreement $\geq$ \acceptanceAgreementPct{}\%, Cohen's $\kappa$ $\geq$ \acceptanceKappaMin{}) were exceeded.
The sample size of \sampleSizeReq{} was determined by Cohen's $\kappa$ with 4 ordinal categories \citep{sim2005kappa}; \humanGradesTotal{} grades were collected, exceeding the minimum.

The final consensus uses 6 graders for the \consensusNoMajority{} no-majority pairs and 5 graders for the remaining \consensusRemainingPairs{} pairs.
The \ngraders{} local graders run at temperature \graderTemperature{} (deterministic); Opus is an API model without strict reproducibility, anchored to human judgment through the \humanGradesTotal{} validated grades.

\subsection{Metrics}

\begin{itemize}[noitemsep]
  \item \textbf{Mean Reciprocal Rank (MRR)}: Reciprocal rank of the first perfectly relevant result (grade 3).
  \item \textbf{Mean Grade}: Average relevance grade on the 0--3 scale.
  \item \textbf{P@1}: Proportion of queries where the top result received grade 3 (perfectly relevant).
  \item \textbf{nDCG@10}: Normalized Discounted Cumulative Gain at rank 10 \citep{jarvelin2002cumulated}.
\end{itemize}

\paragraph{Metric hierarchy for conversation recall.}
This evaluation addresses \emph{single-conversation recall}, meaning the task is to locate a specific past conversation, not \emph{multi-document retrieval} where several relevant documents must be ranked.
The task is analogous to passage retrieval in question answering \citep{voorhees2005trec}, where the goal is to locate a single answer.
In this paper, that answer is usually a resurfaced memory from the same user's own prior work with the agent.

We adopt MRR as the primary metric, following established IR evaluation practice for passage ranking \citep{voorhees2005trec,thakur2021beir}.
MRR measures the reciprocal rank of the first perfectly relevant result, directly quantifying search efficiency: an MRR of 0.50 indicates the first relevant result appears at position $\sim$2 on average.
Empirically, MRR ranges from \vectorMRRmin{} to \keywordMRRmax{} across configurations, indicating that the first perfectly relevant result typically appears within the top 2--3 positions.

Mean Grade quantifies result quality independent of rank.
P@1 isolates top-result precision.
nDCG@10 evaluates ranking quality over the top 10 results; we report @10 following TREC and BEIR (Benchmarking IR) \citep{jarvelin2002cumulated,thakur2021beir} conventions for comparability, though empirically observed examination depth is 2--5 results.
Tables are sorted by MRR (descending), with secondary sorts by Mean Grade, P@1, and nDCG@10.

\subsection{Statistical Methods}

Paired $t$-tests compare per-query mean grades across configurations (each query serves as its own control).
Because the 0--3 grade scale is ordinal and the distribution is skewed, with grades 1 and 3 accounting for \gradeOnePct{}\% and \gradeThreePct{}\% of consensus grades respectively, we report Wilcoxon signed-rank tests alongside the paired $t$-tests as a non-parametric robustness check.
For each of the 4 retrieval mechanisms, we compare the verbatim baseline against each of the 10 distilled configurations (4 single-mechanism $\times$ 1 fusion + 3 multi-field $\times$ 3 fusions = 10), yielding 40 comparisons total; the significance threshold is Bonferroni-corrected to $\alpha$ = 0.05/40 = 0.00125.
Effect sizes are reported as paired Cohen's $d$ ($d_z$: mean difference divided by standard deviation of differences; \citealt{cohen1988statistical}).
95\% confidence intervals for individual configuration means use the normal approximation; 95\% CIs for pairwise mean differences and effect sizes use bootstrap (10{,}000 resamples).

% ============================================================================
% 5. RESULTS
% ============================================================================
\section{Results}
\label{sec:results}

\subsection{Main Results}

\begin{table}[htbp]
\centering
\caption{Headline retrieval results. Full configuration-level results appear in Appendix~\ref{app:full-results-tables}.}
\label{tab:main-results-summary}
\small
\begin{tabular}{p{2.4cm}p{7.4cm}c}
\toprule
Family & Configuration & MRR \\
\midrule
Best verbatim & \bestVerbatimLabel{} & \bestVerbatimMRR{} \\
Best pure distilled & \bestDistilledLabel{} & \bestDistilledMRR{} \\
Best cross-layer & \bestCrossLayerLabel{} & \bestCrossLayerMRR{} \\
\bottomrule
\end{tabular}
\end{table}

\begin{figure}[htbp]
  \centering
  \includegraphics[width=0.85\textwidth]{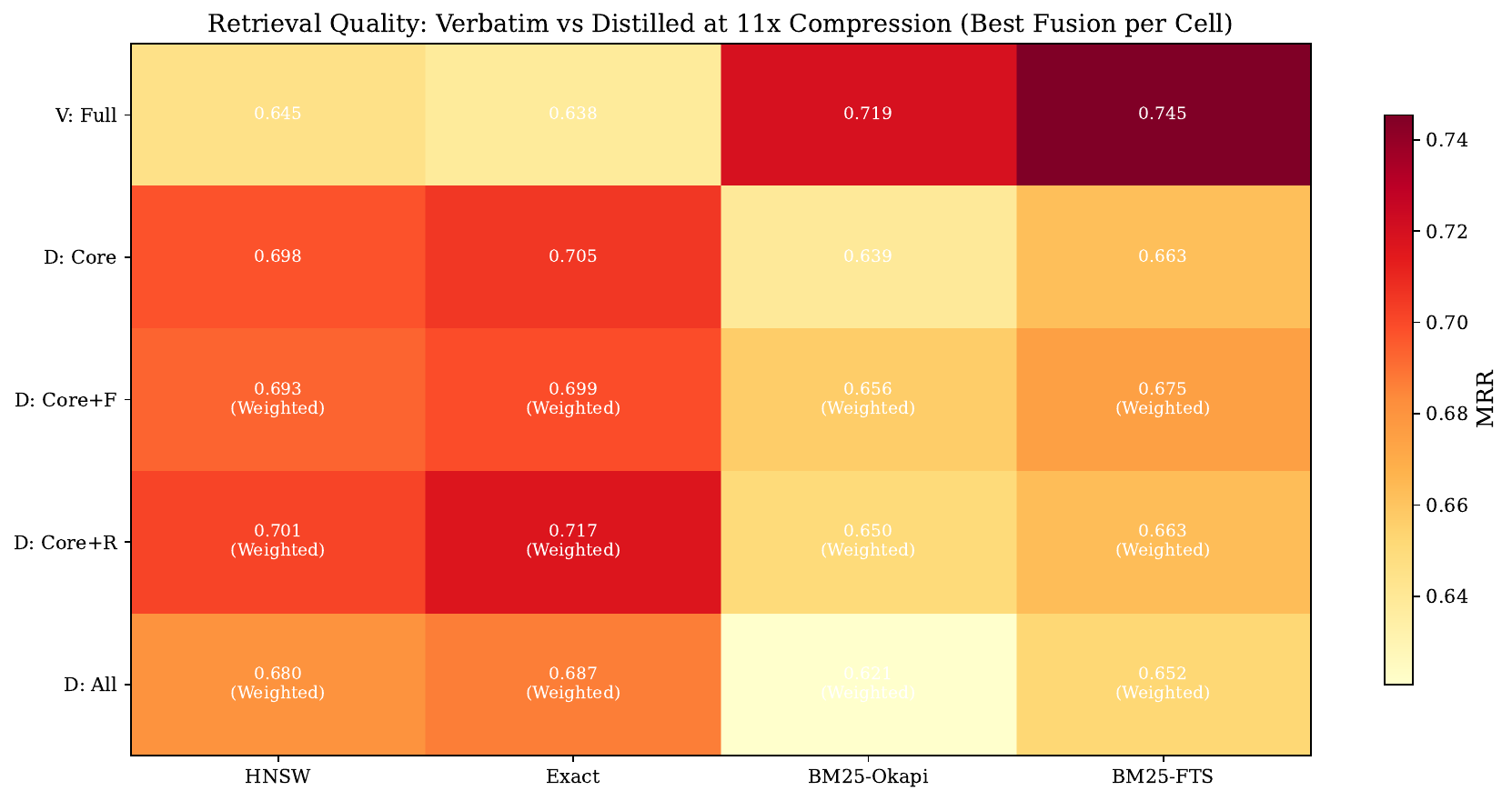}
  \caption{MRR heatmap by mode and retrieval mechanism (best fusion per cell). Darker cells indicate higher MRR.}
  \label{fig:mode-comparison}
\end{figure}

Table~\ref{tab:main-results-summary} and Figure~\ref{fig:mode-comparison} present the headline results. The full ranking of all \numconfigs{} configurations is moved to Appendix~\ref{app:full-results-tables}.

% NOTE: The following narrative must match Table 1 (auto-generated, sorted by MRR descending).
% After data changes, verify the ranking claims against the table. Do NOT hardcode rankings.

\textbf{Best pure distilled.}
The highest-MRR pure distilled configuration is \bestDistilledLabel{} (MRR=\bestDistilledMRR{}), preserving \bestDistilledMRRPct{}\% of the best verbatim MRR (\bestVerbatimLabel{}, MRR=\bestVerbatimMRR{}), \bestDistilledGradePct{}\% of mean grade, \bestDistilledPAtOnePct{}\% of P@1, and \bestDistilledNDCGPct{}\% of nDCG@10.

\textbf{Cross-layer complementarity.}
The highest-MRR configuration overall is a cross-layer setup that fuses BM25 keyword search on verbatim text with HNSW vector search on distilled text (MRR=\bestCrossLayerMRR{}).
It slightly exceeds the best pure verbatim baseline (\bestCrossLayerMRRPct{}\% of verbatim MRR), but it uses both verbatim and distilled text. This is not distilled superiority. It is complementarity.

\textbf{Verbatim baselines.}
Among pure verbatim configurations, BM25-FTS achieves the highest MRR (\verbatimBmftsMRR{}), followed by BM25-Okapi (\verbatimBmokapiMRR{}), HNSW (\verbatimHnswMRR{}), and Exact (\verbatimExactMRR{}).

\textbf{Mechanism-mode interaction.}
Mechanism choice matters more than distillation mode: within any mode, the MRR gap between the best and worst mechanism exceeds the gap between modes for any given mechanism.
The verbatim baselines span MRR \verbatimExactMRR{} (Exact) to \verbatimBmftsMRR{} (BM25-FTS), a range of \spreadverbatim{}.
For mean grade, Full Text / HNSW achieves \verbatimHnswGrade{}; the closest distilled match is \closestDistilledGradeLabel{} at \closestDistilledGrade{} (\closestDistilledGradePct{}\% of verbatim).
Figure~\ref{fig:mode-comparison} makes this plain. The high-scoring cells cluster by mechanism, not by distillation mode.

\subsection{Distillation Quality: Mechanism-Dependent Preservation at \compressionratio{}x Compression}
\label{sec:compression}

Quality preservation depends strongly on the retrieval mechanism.

\textbf{Vector search (HNSW, Exact).}
With HNSW, \hnswnotsig{} of \hnswtotal{} distilled configurations show no statistically significant mean grade degradation after Bonferroni correction ($\alpha$=\bonferronialpha{}); with Exact, \exactnotsig{} of \exacttotal{} are non-significant (Table~\ref{tab:stat-tests}).
The closest distilled match to verbatim HNSW is \closestDistilledGradeLabel{} at \closestDistilledGrade{} (\closestDistilledGradePct{}\% of verbatim mean grade \verbatimHnswGrade{}).
Effect sizes for vector search comparisons range from negligible to small ($|d| \leq 0.25$).
Vector similarity captures semantic content preserved by distillation even after \compressionratiotext{} compression.

\textbf{Keyword search (BM25).}
\bmokapisig{} of \bmokapitotal{} BM25-Okapi and \bmftssig{} of \bmftstotal{} BM25-FTS comparisons show statistically significant degradation after Bonferroni correction.
BM25 effect sizes range from small to medium; \nmediumeffects{} of \totalcomparisons{} comparisons reach the medium threshold ($d \geq 0.5$), with the largest occurring in All Facets configurations ($d$=\maxdcohen{}).
BM25 depends on lexical overlap, and \compressionratiotext{} compression removes the verbatim vocabulary that keyword matching requires.

\begin{figure}[t]
  \centering
  \includegraphics[width=0.85\textwidth]{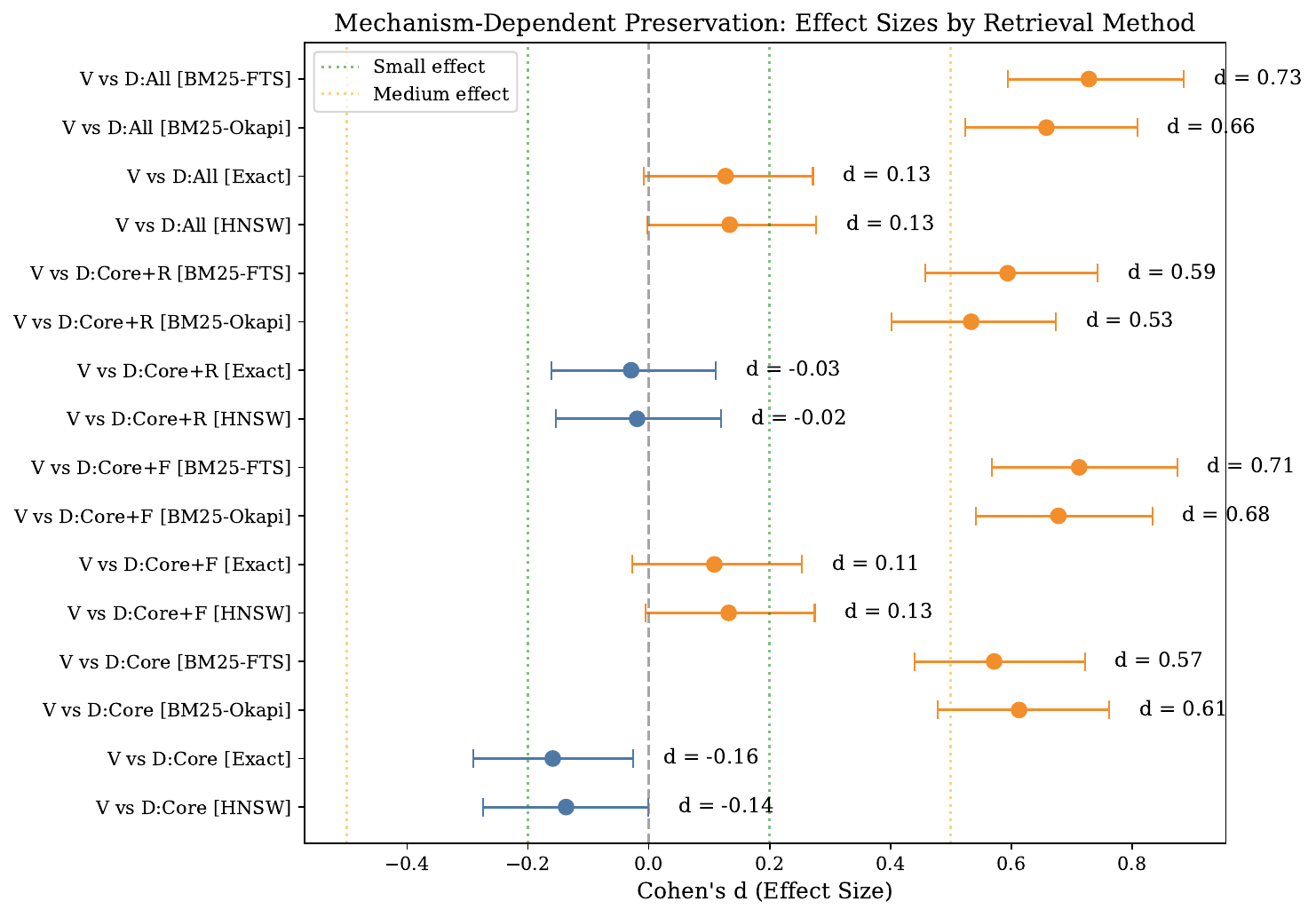}
  \caption{Effect sizes (Cohen's $d$) for per-mechanism comparisons (verbatim vs each distilled mode, averaging across fusion strategies within each mode) with 95\% bootstrap confidence intervals. Grouped by mechanism.}
  \label{fig:effect-sizes}
\end{figure}

\textbf{MRR vs mean grade.}
Mean grade degradation is modest (the closest distilled match preserves \closestDistilledGradePct{}\% of verbatim grade).
The best pure distilled configuration (\bestDistilledLabel{}) achieves MRR=\bestDistilledMRR{}, preserving \bestDistilledMRRPct{}\% of the best verbatim MRR (\bestVerbatimMRR{}).
MRR is sensitive to the rank position of grade-3 results; mean grade averages across all \topk{} results.
Figure~\ref{fig:effect-sizes} says the same thing in another register. Vector-search comparisons sit near zero. BM25 is negative almost throughout, and often by more.

\subsection{Query Type Interaction}

\begin{figure}[htbp]
  \centering
  \includegraphics[width=0.7\textwidth]{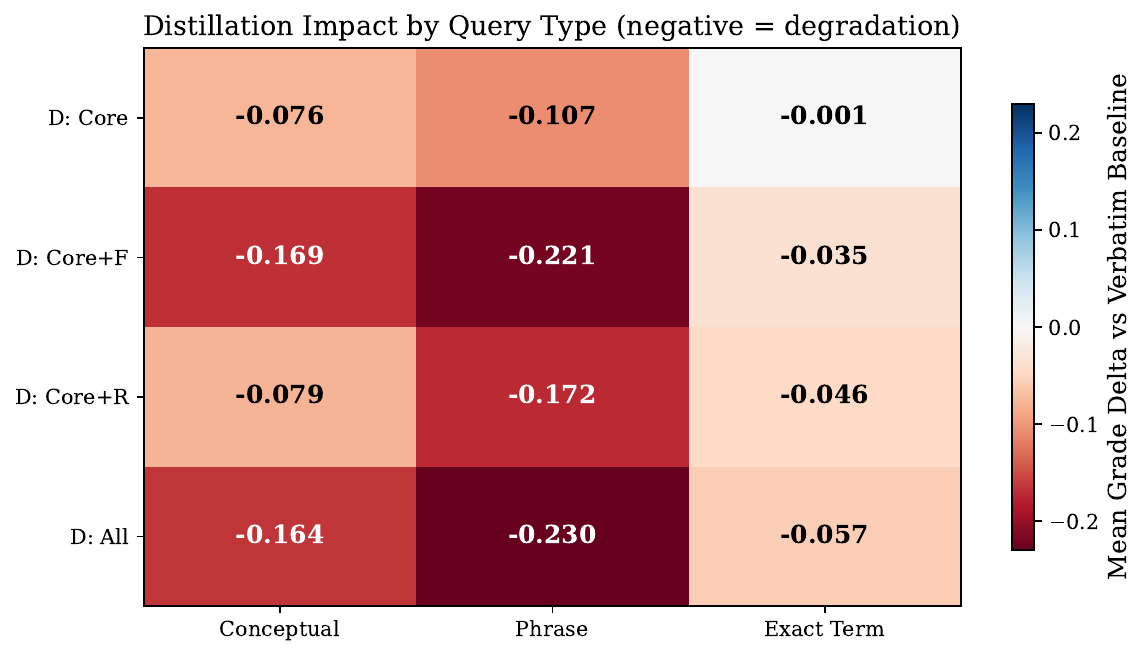}
  \caption{Distillation impact by query type. Heatmap showing mean grade delta relative to pooled verbatim baseline; negative values (red) indicate degradation, positive values (blue) indicate improvement.}
  \label{fig:query-type}
\end{figure}

Not all queries degrade in the same way. Figure~\ref{fig:query-type} shows the main query-type contrasts. The compact comparison table and full query-type breakdown across all \numconfigs{} configurations appear in Appendix~\ref{app:full-results-tables}.
Exact term queries show a reversal: several distilled configurations outperform verbatim on exact terms (e.g., Core+Files / HNSW / Weighted: \qtExactDistillCFHnswWt{} vs Full Text / HNSW: \qtExactVerbatimHnsw{}), consistent with file metadata providing precise identifier matching (see Appendix~\ref{app:best-cases} for individual examples).
Conceptual queries show the largest verbatim advantage (Full Text / HNSW: \qtConceptualVerbatimHnsw{} vs Core+Files / HNSW / RRF: \qtConceptualDistillCFHnswRRF{}), consistent with abstract semantic content being harder to preserve through compression.
Appendix~\ref{app:failure-cases} presents queries where all modes struggle, revealing cases where the target content was either absent from the corpus or discussed with different terminology.

The asymmetry is plain in the heatmap. The largest negative cells sit in conceptual and phrase queries. Exact-term queries stay closer to parity and sometimes flip positive.

\subsection{Grade Distributions}

\begin{figure}[htbp]
  \centering
  \includegraphics[width=0.85\textwidth]{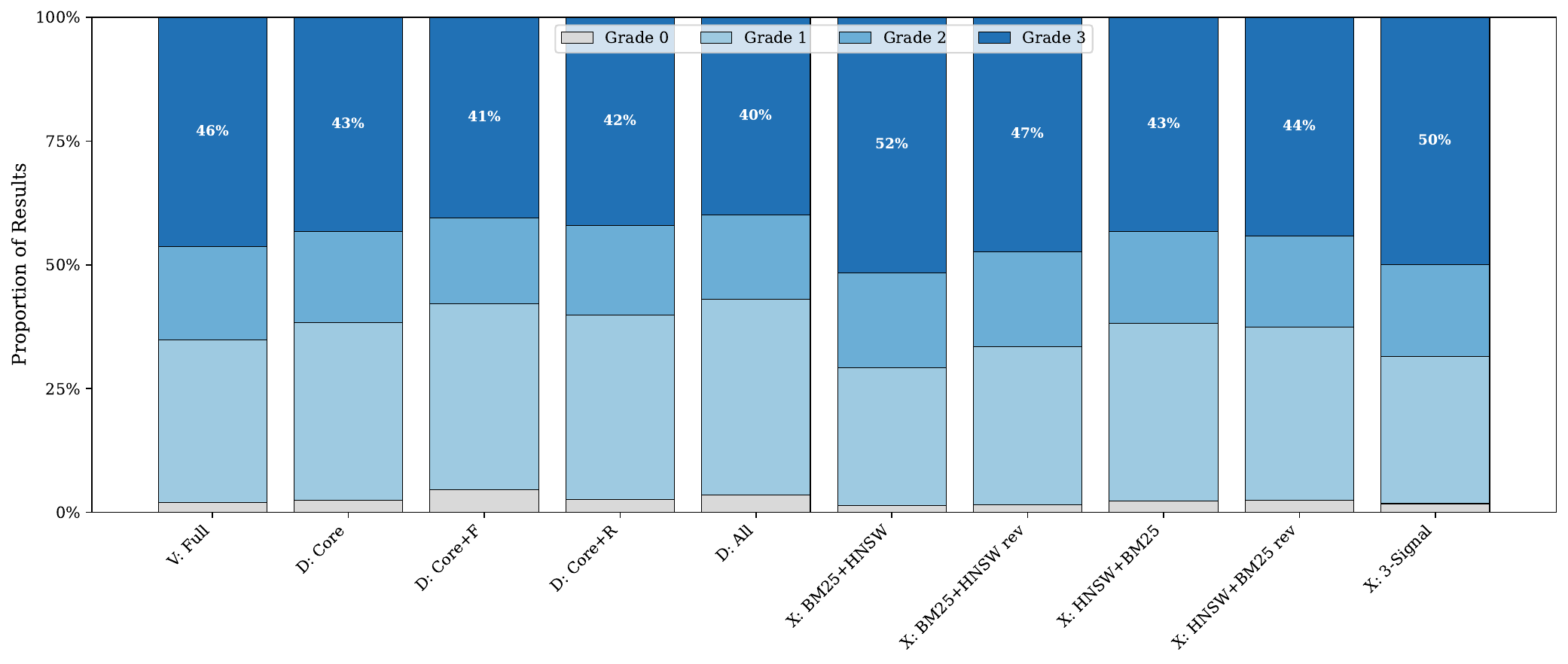}
  \caption{Grade composition by search mode. Bars show the proportion of pooled results at grades 0--3 for each mode. Grade 3 is the largest bucket in every mode, but the overall profiles remain similar, with most mass concentrated in grades 1 and 3.}
  \label{fig:grade-dist}
\end{figure}

That does not mean the whole relevance profile has changed. Figure~\ref{fig:grade-dist} gives the pooled grade composition by mode.

Figure~\ref{fig:grade-dist} shows the grade composition pooled by mode across mechanisms.
Grade 3 is the single largest bucket across modes (\gradeThreePct{}\% of consensus grades overall), while grade 1 remains substantial at \gradeOnePct{}\%.
The relative mix changes only modestly across modes. BM25-heavy settings place slightly more mass away from grade 3 and into the lower grades.
This rules out a simpler story. Distillation is not wrecking the relevance distribution everywhere. The differences are narrower. They show up in specific mechanisms and query types.

\clearpage
\subsection{Inter-Rater Agreement}

\begin{table}[H]
\centering
\caption{Agreement summary. The full pairwise agreement table appears in Appendix~\ref{app:full-results-tables}.}
\label{tab:agreement-summary}
\small
\begin{tabular}{p{5.5cm}p{6.7cm}}
\toprule
Measure & Value \\
\midrule
Fleiss' $\kappa$ across all graders & \fleissk{} (\fleisskinterp{}), over \fleissn{} items \\
Best pairwise $\kappa$ & \bestpairnameA{} / \bestpairnameB{}: \bestpairkappa{} (\bestpairinterp{}) \\
Worst pairwise $\kappa$ & \worstpairnameA{} / \worstpairnameB{}: \worstpairkappa{} (\worstpairinterp{}) \\
\bottomrule
\end{tabular}
\end{table}

\begin{figure}[htbp]
  \centering
  \includegraphics[width=0.7\textwidth]{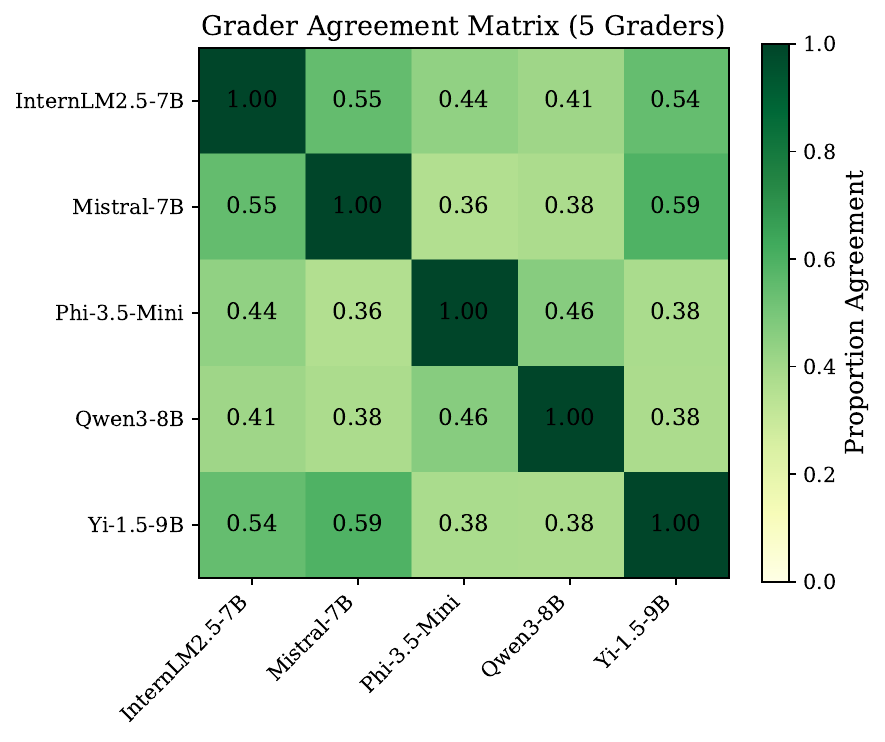}
  \caption{Pairwise proportion agreement heatmap among \ngraders{} LLM graders. Values show the fraction of items where two graders assigned identical grades. Cohen's $\kappa$ values (Table~\ref{tab:agreement-summary} and Appendix~\ref{app:full-results-tables}) are lower due to chance correction.}
  \label{fig:agreement}
\end{figure}

Fleiss' $\kappa$ \citep{fleiss1971measuring} across all \ngraders{} graders (Table~\ref{tab:agreement-summary}) is \fleissk{} (\fleisskinterp{} agreement, \citealt{landis1977measurement}), computed over \fleissn{} items rated by all graders (Figure~\ref{fig:agreement}).
The best pairwise agreement is between \bestpairnameA{} and \bestpairnameB{} ($\kappa$=\bestpairkappa{}, \bestpairinterp{}); the worst is between \worstpairnameA{} and \worstpairnameB{} ($\kappa$=\worstpairkappa{}, \worstpairinterp{}).

This level of agreement is low.
The important point is consistency: all five graders show verbatim modes outperforming distilled modes within BM25, and all five show smaller or non-significant degradation with vector search mechanisms.
Figure~\ref{fig:agreement} makes the structure easier to read than the kappas alone. Exact pairwise agreement is modest, but no grader pair flips the pattern.

The agreement noise is real. It does not erase the main pattern. The next subsection tests that formally.

\subsection{Statistical Tests Summary}

\begin{table}[htbp]
\centering
\caption{Statistical summary by retrieval mechanism. The full comparison table appears in Appendix~\ref{app:full-results-tables}.}
\label{tab:stat-summary}
\small
\begin{tabular}{lcc}
\toprule
Mechanism & Significant & Non-significant \\
\midrule
HNSW & \hnswsig{}/\hnswtotal{} & \hnswnotsig{}/\hnswtotal{} \\
Exact & \exactsig{}/\exacttotal{} & \exactnotsig{}/\exacttotal{} \\
BM25-Okapi & \bmokapisig{}/\bmokapitotal{} & \bmokapinotsig{}/\bmokapitotal{} \\
BM25-FTS & \bmftssig{}/\bmftstotal{} & \bmftsnotsig{}/\bmftstotal{} \\
\bottomrule
\end{tabular}
\end{table}

Table~\ref{tab:stat-summary} presents the \totalcomparisons{} per-mechanism comparisons (4 mechanisms $\times$ 10 distilled configurations each), restricted to pure modes where distilled and verbatim share the same retrieval mechanism. The full row-level comparison table is moved to Appendix~\ref{app:full-results-tables}.

\textbf{Main result:} Of \totalcomparisons{} comparisons, \totalsig{} show statistically significant mean grade degradation after Bonferroni correction ($\alpha$=\bonferronialpha{}).
All \totalnotsig{} non-significant comparisons are in HNSW and Exact; all \totalsig{} significant comparisons are in BM25.

\textbf{Per-mechanism pattern.}
HNSW and Exact account for all \totalnotsig{} non-significant comparisons; BM25-Okapi and BM25-FTS account for all \totalsig{} significant ones.

\textbf{Effect sizes} range from $|d|$=\mindcohen{} to \maxdcohen{}.
Of \totalcomparisons{} comparisons, \nmediumeffects{} reach the medium threshold ($d \geq 0.5$), all in BM25; the largest is $d$=\maxdcohen{}.
The largest effects occur with BM25 keyword search, where lexical overlap loss from compression does the most damage.

Wilcoxon signed-rank tests confirm the same significance pattern as the $t$-tests.
The key finding is simple: distillation quality preservation is mechanism-dependent. With vector search, all \totalnotsig{} configurations achieve non-significant degradation at \compressionratiotext{} compression; with keyword search, all \totalsig{} configurations show significant degradation, with effect sizes ranging from small to medium ($|d|$=\mindcohen{}--\maxdcohen{}).

\subsection{Coverage Analysis: Verbatim vs Distilled Result Pools}
\label{sec:coverage}

\begin{figure}[htbp]
  \centering
  \includegraphics[width=0.95\textwidth]{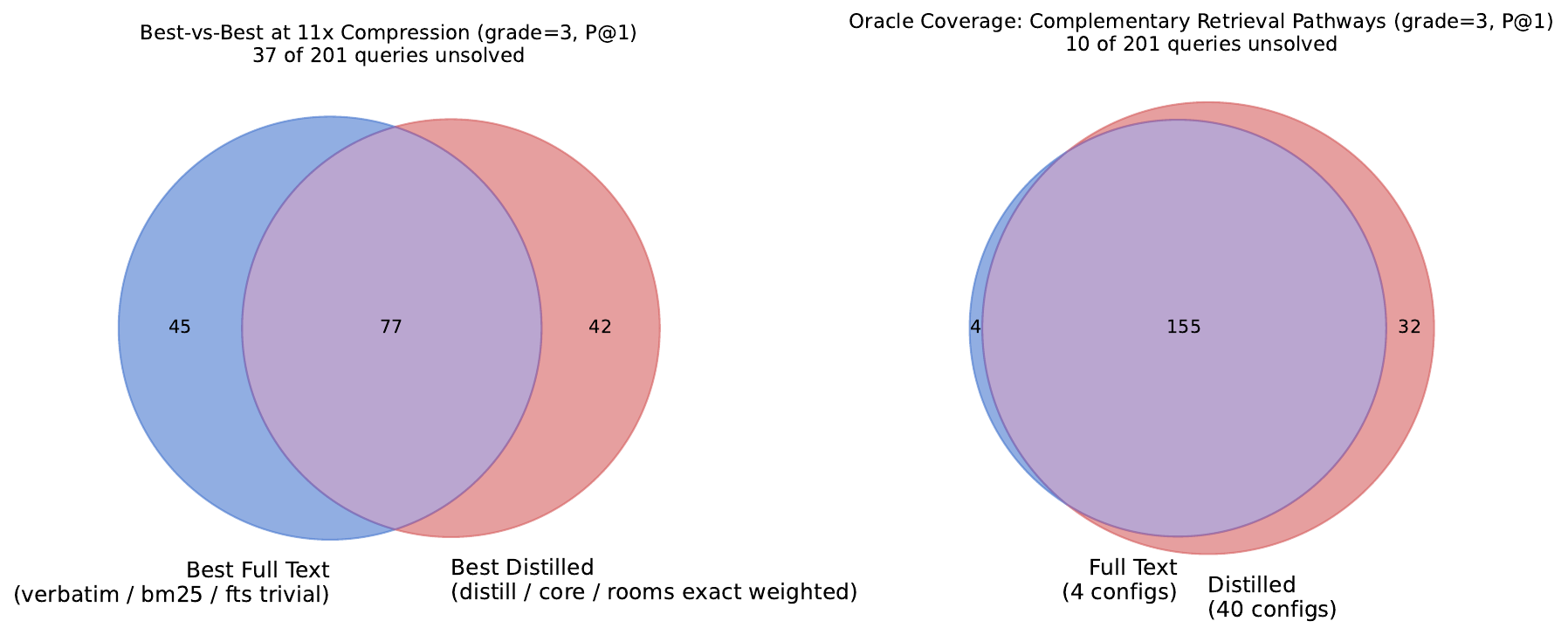}
  \caption{Query coverage overlap between Full Text and Distilled configurations (grade-3, P@1). Left: best single config from each family, showing near parity at \compressionratiotext{} compression. Right: oracle over all configs per family, showing complementary retrieval pathways.}
  \label{fig:venn-coverage}
\end{figure}

The ranking metrics already hint at complementarity. Figure~\ref{fig:venn-coverage} shows it directly at the query level.

Figure~\ref{fig:venn-coverage} partitions the \numqueries{} queries by which configurations solve them at P@1 (grade=3).

\textbf{Best-vs-best parity.}
The left panel compares the single best configuration from each family.
The best Full Text configuration solves \bestVSolves{} queries; the best Distilled configuration solves \bestDSolves{} queries.
The overlap is \bestOverlap{} queries; \bestVOnly{} are solved only by Full Text, \bestDOnly{} only by Distilled.
At the level of the best individual configuration, distillation matches verbatim retrieval at \compressionratiotext{} compression, with near-symmetric exclusive coverage.

\textbf{Oracle complementarity.}
The right panel shows oracle coverage using all configurations per family (\oracleVConfigs{} Full Text, \oracleDConfigs{} Distilled).
The union solves \oracleUnion{} of \numqueries{} queries (\oracleUnionPct{}\%), with only \oracleNeither{} unsolvable.
The \oracleDOnly{} distill-only queries span all three query types (\oracleDOnlyConceptual{} conceptual, \oracleDOnlyPhrase{} phrase, \oracleDOnlyExact{} exact term). Structured distillation is creating retrieval pathways through exchange cores, room assignments, and file metadata that the verbatim configurations miss.
The asymmetry (\oracleDOnly{} distill-only vs \oracleVOnly{} verbatim-only) is partly a configuration count effect (\oracleDConfigs{} vs \oracleVConfigs{}) and does not indicate that distilled text is inherently more searchable; the fair best-vs-best comparison in the left panel controls for this.

That complementary coverage raises a natural question. Can the verbatim text, which is already stored for display, serve as a second-stage reranking signal for distilled results?

\paragraph{Post-hoc reranking experiment.}
We tested this by applying BM25 snippet reranking to the top-\topk{} distilled candidates: each candidate's verbatim snippet is scored against the query via BM25-Okapi and blended with the original retrieval score ($\lambda$=\rerankAlpha{} original, \rerankBeta{} BM25).
The best distilled configuration after reranking (\rerankTargetConfig{}) achieves \rerankDistillReranked{}/\rerankTotalQueries{} queries at P@1, compared to \rerankVerbatimBaseline{}/\rerankTotalQueries{} for Full Text / BM25-FTS.
The gap narrows from \rerankGapBefore{} queries (\rerankDistillOriginal{} vs \rerankVerbatimBaseline{} without reranking) to \rerankGapAfter{} queries (\rerankDistillReranked{} vs \rerankVerbatimBaseline{}), but does not close.

The reranker promotes \rerankGained{} queries to P@1 by moving up candidates ranked 2--7 when BM25 snippet matching identifies the right result. It also demotes \rerankLost{} queries from P@1 when a wrong candidate has higher lexical overlap in the wrong context.
In every loss, the promoted candidate's snippet contains more query terms than the correct result's snippet, but those terms appear in an irrelevant discussion.
The same mechanism causes both behaviors. Rewarding verbatim term overlap creates the \rerankGained{} wins and the \rerankLost{} failures.

We tested whether any query-time feature could gate the reranker to prevent losses: original retrieval score, score margin between rank 1 and rank 2, BM25 score of the rank-1 result, and query term overlap fraction.
None separates gains from losses.
The distributions overlap. Losses have \emph{higher} median BM25 scores and \emph{more} term overlap than gains, which is the opposite of what a protective threshold would need.
The best single-feature gate achieves \rerankBestGate{}/\rerankTotalQueries{}, worse than both the ungated reranker (\rerankDistillReranked{}) and the verbatim baseline (\rerankVerbatimBaseline{}).

The features we tested do not separate gains from losses. The missing signal may not be textual.
Online learning from user interaction, such as click-through rates, dwell time, and query reformulation, could supply the missing signal. That requires a deployed system with real users, so it sits outside this offline evaluation.

% ============================================================================
% 6. DISCUSSION
% ============================================================================
\section{Discussion}

\subsection{Context Compression, Not Search Optimization}

The search evaluation stands in for a simpler question: does distillation preserve the information that matters?
At \compressionratiotext{} compression, thousands of conversation exchanges fit within a single agent prompt, enabling persistent memory without context overflow.

The distilled text is never shown to the user.
It functions as a routing index. Distillation determines which conversations surface and in what rank order, but the search result the user reads is the verbatim exchange that produced it.
Index representation and display representation are separate layers.
If the distilled index routes correctly, the user sees the original conversation text; compression artifacts in the distilled form are invisible.

Each distilled object carries a \texttt{conversation\_id} and ply-range back-reference to its source exchange.
A search hit returns the matched verbatim snippet directly. The user sees the exact user and assistant turns from the original conversation, not a summary or reconstruction.
The distilled layer determines \emph{which} exchange to surface; the verbatim layer determines \emph{what} the user reads.

This two-tier separation gives a concrete architectural alternative to compaction for personalized agent memory.
In current agent systems, when the context window fills, the agent summarizes its own history and discards the original. That is lossy compression applied iteratively, so the information loss compounds over long sessions.
The verbatim conversation, meanwhile, is already persisted on the client.
The distillation results point to a different approach. The agent carries only the distilled form as working memory, at \compressionratiotext{} the token cost, while the complete verbatim text remains on the client as ground truth.
When the distilled form is sufficient, which is the common case given that all \totalnotsig{} vector search configurations show no significant degradation, the agent operates at a fraction of the context cost.
When it is not, because the user needs exact wording or the distilled form dropped a specific detail, verbatim retrieval from local storage resolves it.
We do not evaluate agent task performance under distilled-only context here; that direct comparison remains future work.
What the search evaluation does establish is simpler. Structured distillation preserves enough information for retrieval at \compressionratiotext{} compression. Without that, the whole idea falls apart.

The search evaluation is, in fact, a harder test than what live context management requires.
Search has two failure points: the distilled text must \emph{rank} correctly against competing candidates, and it must \emph{contain} the relevant information.
Ranking is where our degradation comes from, particularly for BM25, where lexical compression removes the vocabulary that keyword matching depends on.
In live context management, there is no ranking.
Completed exchanges are loaded sequentially, in conversation order; the only question is whether the distilled form preserves enough information for the agent to understand what occurred.
The grade evaluation measures information preservation independent of rank position. It shows small to medium effect sizes ($|d|$=\mindcohen{}--\maxdcohen{}), with the largest confined to BM25 keyword search.
The retrieval degradation that is our primary limitation is less central in the sequential-loading case; the evidence for information preservation still matters there.

\subsection{The Asymmetric Memory Problem}

The current state of human-agent conversation is asymmetric: the human retains fragments of past interactions through recognition memory, while the agent starts each session with no history.
This asymmetry degrades conversation quality over time. The agent re-asks questions that were answered weeks ago, re-proposes solutions that were tried and rejected, and cannot build on prior decisions.

Distilled memory addresses this by giving both parties access to the same compressed history.
The agent can reference past decisions (``In session 47, we chose connection pooling over per-request connections because of the timeout issue''), and the human can verify against their own recognition memory.

Git is a distillation system for code: the commit message is a distilled object summarizing what changed; the file tree provides spatial organization; \texttt{git log --follow} walks into a room.
A project with 50{,}000 commits remains navigable because developers interact with distilled summaries (commit messages), not raw diffs.
We build the same thing for conversations.

\subsection{Complementary Retrieval Signals}

The coverage analysis (Section~\ref{sec:coverage}) shows that verbatim and distilled result pools are partially complementary: \bestDOnly{} queries are solved only by distilled configurations while \bestVOnly{} are solved only by verbatim.
Distillation does not just lose signal; it creates a second signal that verbatim search misses.
The cross-layer search modes (Section~\ref{sec:results}) confirm this: fusing BM25 on verbatim text with HNSW on distilled text achieves the highest MRR overall (\bestCrossLayerMRR{}), slightly exceeding the best pure verbatim configuration (\bestVerbatimMRR{}).
Distillation is not just a degraded copy of the original signal. It produces a complementary signal that improves retrieval when fused with verbatim keyword search.

\subsection{Mechanism Dominance Within Pure Modes}

Within pure modes, the dominant factor is retrieval mechanism, not distillation mode.
Vector similarity (HNSW, Exact) captures semantic content preserved through distillation: the surviving vocabulary principle ensures that query-relevant terms survive compression (\queryOverlapRetention{} of query vocabulary retained), and embedding similarity detects this preserved meaning.
BM25 depends on lexical overlap, and \compressionratiotext{} compression removes exchange-specific vocabulary (only \vocabSurvivalRate{} of top-\vocabSurvivalK{} IDF tokens survive) that keyword matching requires.
HNSW: \hnswnotsig{} of \hnswtotal{} comparisons are non-significant after Bonferroni correction; Exact: \exactnotsig{} of \exacttotal{}.
All \hnswnotsig{} HNSW and all \exactnotsig{} Exact comparisons are non-significant regardless of fusion strategy; significant degradation is confined entirely to BM25 keyword search.

\subsection{Per-Grader Consistency}

The inter-rater agreement ($\kappa$=\fleissk{}) is low by standard benchmarks; human assessors in TREC evaluations achieve $\kappa$=0.35--0.65 \citep{voorhees1998variations}.
The task is genuinely difficult: assessing whether a fragment of a software engineering conversation is ``relevant'' to a short query requires domain understanding that 7--12B parameter models may lack.

Despite this disagreement on absolute scores, the mechanism-dependent preservation finding is directionally consistent across all five graders: all five show verbatim modes outperforming distilled modes within BM25, and all five show smaller or non-significant degradation with vector search.
The length-bias hypothesis does not hold up. If short distilled snippets were being rewarded across the board, all distilled modes should benefit uniformly across mechanisms. They do not. BM25 degrades much more than HNSW.

% ============================================================================
% 7. LIMITATIONS
% ============================================================================
\section{Limitations}

We present ten limitations, ordered by severity.

\begin{enumerate}

\item \textbf{Low inter-rater agreement} ($\kappa$=\fleissk{}).
The five LLM graders barely agree beyond chance.
Majority voting mitigates but does not eliminate this noise.
All quantitative results should be interpreted with this noise floor in mind.

\item \textbf{Mechanism-dependent degradation in pure-mode comparisons.}
\totalsig{} of \totalcomparisons{} pure-mode per-mechanism comparisons show statistically significant mean grade degradation after Bonferroni correction ($\alpha$=\bonferronialpha{}).
All significant comparisons are BM25 keyword search; all \totalnotsig{} vector search comparisons are non-significant.
The best pure distilled configuration preserves \bestDistilledMRRPct{}\% of the best verbatim MRR (\bestDistilledMRR{} vs \bestVerbatimMRR{}); cross-layer configurations slightly exceed verbatim but rely on fusing verbatim and distilled signals.
With vector search, compression is mostly fine. With BM25, it costs you.

\item \textbf{Single-user setting.}
All data comes from one developer's conversations across 6 projects.
This is intentional: the paper studies personalized memory for one user's long-running agent history.
What it does \emph{not} establish is transfer across different users, domains, conversation styles, or languages, which may produce different compression ratios and different preservation profiles.

\item \textbf{Limited human evaluation.}
The primary grading was performed by \graderParamRange{} parameter local LLMs, with a calibrated Claude Opus grader for the \consensusNoMajority{} no-majority pairs.
Human validation (\humanGradesTotal{} grades, Section~\ref{sec:human-validation}) targets the Opus grader's accuracy on the decisive stratum, not a random sample of all grades.

\item \textbf{No head-to-head comparison.}
We did not benchmark against Letta, Mem0, Zep, or other memory systems.
Our evaluation asks whether a personalized distilled memory layer is viable within this system, not whether it dominates other memory architectures head-to-head.

\item \textbf{Effect sizes range from small to medium.}
The \totalcomparisons{} pairwise effect sizes range from $|d|$=\mindcohen{} to \maxdcohen{}.
\nmediumeffects{} of \totalcomparisons{} comparisons reach the medium threshold ($d \geq 0.5$), all in BM25; the remaining are small.

\item \textbf{Spatial navigation not evaluated.}
Rooms are evaluated only as retrieval metadata.
The spatial navigation interface (walking from room to room, exploring connections) is not evaluated.
What we measure here is rooms-as-metadata, not rooms-as-UI.
The user experience of navigating a memory palace remains untested.

\item \textbf{Distillation quality depends on the LLM.}
All \numdistilledobjects{} evaluated palace objects were produced by Claude Haiku 4.5 \citep{anthropic2025haiku45}.
Different models at different capability levels may produce different compression ratios and quality preservation.

\item \textbf{Length-bias confound in LLM grading.}
Distilled snippets average $\sim$\distilledAvgChars{} characters while verbatim snippets average $\sim$\verbatimTruncChars{} characters; the \graderParamRange{} graders may exhibit systematic bias toward shorter or longer text.
Short-text bias would inflate distilled mode scores uniformly across mechanisms, but BM25 shows consistently larger degradation than HNSW.
That pattern is hard to square with uniform short-text bias. It is better explained by real differences in how retrieval mechanisms handle compressed text.

\item \textbf{Recall interface not implemented.}
The conversational recall pattern, where a user describes a memory vaguely and the system proposes candidates for recognition, is described in the design but remains future work.

\end{enumerate}

% ============================================================================
% 8. CONCLUSION
% ============================================================================
\section{Conclusion}

Structured distillation compresses conversation exchanges by \compressionratiotext{} (\avgverbatimtokens{} $\to$ \avgdistilledtokens{} tokens).
The best pure distilled configuration (\bestDistilledLabel{}) preserves \bestDistilledMRRPct{}\% of the best verbatim MRR (\bestDistilledMRR{} vs \bestVerbatimMRR{}).
Cross-layer configurations that fuse verbatim keyword search with distilled vector search slightly exceed the best pure verbatim baseline (MRR \bestCrossLayerMRR{}). The two representations help in different ways.

Within pure single-mechanism comparisons, the mechanism matters. \hnswnotsig{} of \hnswtotal{} HNSW and \exactnotsig{} of \exacttotal{} Exact distilled configurations show no statistically significant mean grade degradation ($\alpha$=\bonferronialpha{}), while \bmokapisig{} of \bmokapitotal{} BM25-Okapi and \bmftssig{} of \bmftstotal{} BM25-FTS do degrade significantly.
Effect sizes range from $|d|$=\mindcohen{} to \maxdcohen{}; \nmediumeffects{} of \totalcomparisons{} comparisons reach the medium threshold, all in BM25.

An agent carrying 1{,}000 distilled exchanges in $\sim$39{,}000 tokens has persistent memory without context overflow. The verbatim alternative would require $\sim$407{,}000 tokens.

The point of the system is personalized memory. One user's ongoing work with an agent can be distilled into a compact retrieval layer that remains useful later, while the verbatim source stays available for drill-down. Distilled objects handle fast recognition. Verbatim text handles full recall.

Future work includes evaluating the hybrid search modes defined in Section 3.4 (same-layer keyword + semantic fusion), cross-user validation across different users and domains, and a conversational recall interface in which users describe memories vaguely and the system proposes candidates for recognition-based selection. Post-hoc snippet BM25 reranking narrows the gap (\rerankDistillReranked{}/\rerankTotalQueries{} vs \rerankVerbatimBaseline{}/\rerankTotalQueries{} at P@1 for the best configurations) but does not close it: the BM25 signal that promotes correct results also demotes them, and the static query-time features we tested do not separate the two cases (Section~\ref{sec:coverage}).

We release the implementation and analysis pipeline as open-source software \citep{searchat2026}.\footnote{\url{https://github.com/Process-Point-Technologies-Corporation/searchat}} Because the corpus consists of one user's private conversation history, the underlying data, evaluation queries, and grading outputs are not publicly released. The repository includes the software, configuration files, and analysis scripts needed to run the same methodology on a user's own data.

% ============================================================================
% REFERENCES
% ============================================================================
\bibliographystyle{plainnat}
\bibliography{references}

% ============================================================================
% APPENDICES
% ============================================================================
\appendix

\section{Full Results Tables}
\label{app:full-results-tables}

The main text reports compact summaries. The full generated tables are included here for completeness.

\subsection{All Configuration Results}
% Table 1: Configuration Results (107 configs)
\begingroup
\footnotesize
\setlength{\tabcolsep}{4pt}
\begin{longtable}{lllrrrrr}
\caption{Retrieval quality across \numconfigs{} configurations: verbatim baselines vs.\ distilled modes at \compressionratiotext{} compression (\numqueries{} queries, 0--3 relevance scale). Sorted by MRR descending. V = Full Text, D = Distill, F = Files, R = Rooms. Add = Additive, Wt = Weighted.}
\label{tab:main-results} \\
\toprule
Mode & Mech & Fus & MRR & Grade & 95\% CI & P@1 & nDCG \\
\midrule
\endfirsthead
\multicolumn{8}{l}{\small\itshape Table~\ref{tab:main-results} continued} \\
\toprule
Mode & Mech & Fus & MRR & Grade & 95\% CI & P@1 & nDCG \\
\midrule
\endhead
\midrule
\multicolumn{8}{r}{\small\itshape Continued on next page} \\
\endfoot
\bottomrule
\\[-1.8ex]
\multicolumn{8}{l}{$^\dagger$ Best distilled configuration per mechanism.} \\
\endlastfoot
Cross: BM25(V)+HNSW(D) & BM25-FTS + HNSW & CMZ & 0.759 & 2.237 & [2.20, 2.28] & 0.622 & 0.667 \\
Cross: BM25(V)+HNSW(D) & BM25-FTS + HNSW & W70 & 0.758 & 2.261 & [2.22, 2.30] & 0.622 & 0.675 \\
Cross: BM25(V)+HNSW(D) & BM25-FTS + HNSW & W90 & 0.757 & 2.235 & [2.20, 2.27] & 0.632 & 0.667 \\
Cross: BM25(V)+HNSW(D) & BM25-Okapi + HNSW & W70 & 0.754 & 2.182 & [2.14, 2.22] & 0.642 & 0.650 \\
Cross: BM25(V)+HNSW(D) & BM25-FTS + HNSW & W95 & 0.751 & 2.234 & [2.20, 2.27] & 0.617 & 0.666 \\
Cross: BM25(V)+HNSW(D) & BM25-FTS + HNSW & Add & 0.750 & 2.254 & [2.22, 2.29] & 0.612 & 0.671 \\
Cross: BM25(V)+HNSW(D) & BM25-FTS + HNSW & W50 & 0.750 & 2.254 & [2.22, 2.29] & 0.612 & 0.671 \\
Cross: 3-Signal & BM25-FTS×2 + HNSW & W90 & 0.749 & 2.232 & [2.19, 2.27] & 0.617 & 0.665 \\
Cross: BM25(V)+HNSW(D) & BM25-FTS + HNSW & RRF & 0.748 & 2.236 & [2.20, 2.28] & 0.612 & 0.664 \\
Cross: BM25(V)+HNSW(D) & BM25-Okapi + HNSW & W80 & 0.748 & 2.181 & [2.14, 2.22] & 0.632 & 0.650 \\
Cross: BM25(V)+HNSW(D) & BM25-FTS + HNSW & Max & 0.748 & 2.210 & [2.17, 2.25] & 0.612 & 0.657 \\
Cross: 3-Signal & BM25-FTS×2 + HNSW & W95 & 0.747 & 2.229 & [2.19, 2.27] & 0.612 & 0.664 \\
V & BM25-FTS & --- & 0.745 & 2.225 & [2.19, 2.26] & 0.610 & 0.661 \\
Cross: 3-Signal & BM25-Okapi×2 + HNSW & W70 & 0.745 & 2.147 & [2.11, 2.19] & 0.632 & 0.638 \\
Cross: BM25(V)+HNSW(D) & BM25-Okapi + HNSW & CMZ & 0.745 & 2.193 & [2.15, 2.23] & 0.622 & 0.652 \\
Cross: BM25(V)+HNSW(D) & BM25-FTS + HNSW & W80 & 0.743 & 2.246 & [2.21, 2.28] & 0.602 & 0.670 \\
Cross: 3-Signal & BM25-Okapi×2 + HNSW & W80 & 0.743 & 2.150 & [2.11, 2.19] & 0.632 & 0.639 \\
Cross: 3-Signal & BM25-FTS×2 + HNSW & Max & 0.742 & 2.189 & [2.15, 2.23] & 0.612 & 0.648 \\
Cross: 3-Signal & BM25-FTS×2 + HNSW & W70 & 0.742 & 2.224 & [2.18, 2.26] & 0.602 & 0.662 \\
Cross: 3-Signal & BM25-FTS×2 + HNSW & W80 & 0.740 & 2.236 & [2.20, 2.28] & 0.597 & 0.665 \\
Cross: BM25(V)+HNSW(D) & BM25-Okapi + HNSW & RRF & 0.738 & 2.186 & [2.15, 2.23] & 0.617 & 0.647 \\
Cross: BM25(V)+HNSW(D) & BM25-Okapi + HNSW & Add & 0.733 & 2.189 & [2.15, 2.23] & 0.607 & 0.650 \\
Cross: BM25(V)+HNSW(D) & BM25-Okapi + HNSW & W50 & 0.733 & 2.189 & [2.15, 2.23] & 0.607 & 0.650 \\
Cross: 3-Signal & BM25-FTS×2 + HNSW & W50 & 0.732 & 2.229 & [2.19, 2.27] & 0.577 & 0.661 \\
Cross: 3-Signal & BM25-Okapi×2 + HNSW & W90 & 0.729 & 2.155 & [2.11, 2.19] & 0.607 & 0.637 \\
Cross: 3-Signal & BM25-Okapi×2 + HNSW & Max & 0.728 & 2.117 & [2.08, 2.16] & 0.592 & 0.622 \\
Cross: BM25(V)+HNSW(D) & BM25-Okapi + HNSW & W90 & 0.727 & 2.167 & [2.13, 2.21] & 0.602 & 0.641 \\
Cross: BM25(V)+HNSW(D) & BM25-Okapi + HNSW & W95 & 0.727 & 2.157 & [2.12, 2.20] & 0.602 & 0.638 \\
Cross: BM25(V)+HNSW(D) & BM25-Okapi + HNSW & Max & 0.726 & 2.159 & [2.12, 2.20] & 0.592 & 0.637 \\
Cross: 3-Signal & BM25-Okapi×2 + HNSW & W50 & 0.723 & 2.142 & [2.10, 2.18] & 0.577 & 0.631 \\
Cross: 3-Signal & BM25-Okapi×2 + HNSW & W95 & 0.721 & 2.152 & [2.11, 2.19] & 0.592 & 0.636 \\
V & BM25-Okapi & --- & 0.719 & 2.149 & [2.11, 2.19] & 0.592 & 0.632 \\
D Core+R & Exact & Wt & 0.717$^\dagger$ & 2.077 & [2.04, 2.12] & 0.592 & 0.603 \\
Cross: BM25(V)+HNSW(D) rev & HNSW + BM25-FTS & W70 & 0.717 & 2.167 & [2.13, 2.21] & 0.567 & 0.639 \\
Cross: 3-Signal & BM25-FTS×2 + HNSW & Add & 0.716 & 2.187 & [2.15, 2.23] & 0.557 & 0.645 \\
Cross: 3-Signal & BM25-FTS×2 + HNSW & CMZ & 0.711 & 2.161 & [2.12, 2.20] & 0.567 & 0.636 \\
Cross: BM25(V)+HNSW(D) rev & HNSW + BM25-FTS & W95 & 0.709 & 2.088 & [2.05, 2.13] & 0.577 & 0.608 \\
Cross: BM25(V)+HNSW(D) rev & HNSW + BM25-FTS & W80 & 0.709 & 2.130 & [2.09, 2.17] & 0.567 & 0.623 \\
Cross: BM25(V)+HNSW(D) rev & HNSW + BM25-FTS & W90 & 0.708 & 2.105 & [2.07, 2.15] & 0.577 & 0.614 \\
D Core & Exact & --- & 0.705 & 2.079 & [2.04, 2.12] & 0.577 & 0.600 \\
Cross: 3-Signal & BM25-FTS×2 + HNSW & RRF & 0.705 & 2.147 & [2.11, 2.19] & 0.557 & 0.631 \\
D Core+R & HNSW & Wt & 0.701$^\dagger$ & 2.071 & [2.03, 2.11] & 0.567 & 0.597 \\
Cross: 3-Signal & BM25-Okapi×2 + HNSW & Add & 0.700 & 2.129 & [2.09, 2.17] & 0.547 & 0.621 \\
D Core+F & Exact & Wt & 0.699 & 2.049 & [2.01, 2.09] & 0.562 & 0.598 \\
D Core & HNSW & --- & 0.698 & 2.076 & [2.04, 2.12] & 0.568 & 0.594 \\
Cross: 3-Signal & BM25-Okapi×2 + HNSW & CMZ & 0.694 & 2.092 & [2.05, 2.13] & 0.552 & 0.609 \\
D Core+F & HNSW & Wt & 0.693 & 2.030 & [1.99, 2.07] & 0.557 & 0.590 \\
Cross: 3-Signal & BM25-Okapi×2 + HNSW & RRF & 0.693 & 2.093 & [2.05, 2.13] & 0.547 & 0.608 \\
Cross: HNSW(V)+BM25(D) rev & BM25-FTS + HNSW & W70 & 0.687 & 2.059 & [2.02, 2.10] & 0.557 & 0.594 \\
D All & Exact & Wt & 0.687 & 2.031 & [1.99, 2.07] & 0.557 & 0.586 \\
D All & HNSW & Wt & 0.680 & 2.022 & [1.98, 2.06] & 0.547 & 0.582 \\
D Core+F & BM25-FTS & Wt & 0.675$^\dagger$ & 2.025 & [1.98, 2.07] & 0.540 & 0.581 \\
Cross: HNSW(V)+BM25(D) rev & BM25-FTS + HNSW & W50 & 0.669 & 2.069 & [2.03, 2.11] & 0.538 & 0.595 \\
Cross: HNSW(V)+BM25(D) rev & BM25-FTS + HNSW & W80 & 0.669 & 2.036 & [2.00, 2.08] & 0.527 & 0.584 \\
Cross: HNSW(V)+BM25(D) rev & BM25-FTS + HNSW & W95 & 0.667 & 2.025 & [1.98, 2.07] & 0.522 & 0.579 \\
Cross: HNSW(V)+BM25(D) & HNSW + BM25-FTS & RRF & 0.664 & 2.060 & [2.02, 2.10] & 0.527 & 0.589 \\
D Core+R & BM25-FTS & Wt & 0.663 & 2.041 & [2.00, 2.08] & 0.520 & 0.583 \\
D Core & BM25-FTS & --- & 0.663 & 2.028 & [1.99, 2.07] & 0.520 & 0.571 \\
Cross: HNSW(V)+BM25(D) & HNSW + BM25-FTS & Add & 0.660 & 2.065 & [2.02, 2.11] & 0.522 & 0.593 \\
Cross: HNSW(V)+BM25(D) & HNSW + BM25-FTS & W50 & 0.660 & 2.065 & [2.02, 2.11] & 0.522 & 0.593 \\
Cross: HNSW(V)+BM25(D) rev & BM25-FTS + HNSW & W90 & 0.660 & 2.025 & [1.98, 2.07] & 0.507 & 0.580 \\
Cross: HNSW(V)+BM25(D) & HNSW + BM25-FTS & CMZ & 0.659 & 2.072 & [2.03, 2.11] & 0.517 & 0.594 \\
D Core+F & BM25-Okapi & Wt & 0.656$^\dagger$ & 1.926 & [1.88, 1.97] & 0.520 & 0.549 \\
Cross: HNSW(V)+BM25(D) & HNSW + BM25-Okapi & W70 & 0.656 & 2.017 & [1.98, 2.06] & 0.532 & 0.574 \\
D Core+R & Exact & Add & 0.656 & 1.998 & [1.96, 2.04] & 0.498 & 0.569 \\
Cross: HNSW(V)+BM25(D) & HNSW + BM25-FTS & W80 & 0.653 & 2.018 & [1.98, 2.06] & 0.512 & 0.576 \\
D All & BM25-FTS & Wt & 0.652 & 1.989 & [1.95, 2.03] & 0.515 & 0.564 \\
D Core+R & HNSW & Add & 0.651 & 1.992 & [1.95, 2.03] & 0.502 & 0.565 \\
Cross: HNSW(V)+BM25(D) & HNSW + BM25-Okapi & W80 & 0.651 & 2.004 & [1.96, 2.05] & 0.512 & 0.571 \\
D Core+R & BM25-Okapi & Wt & 0.650 & 1.960 & [1.92, 2.00] & 0.515 & 0.556 \\
Cross: HNSW(V)+BM25(D) & HNSW + BM25-FTS & Max & 0.649 & 2.051 & [2.01, 2.09] & 0.498 & 0.587 \\
Cross: HNSW(V)+BM25(D) & HNSW + BM25-Okapi & Max & 0.649 & 2.010 & [1.97, 2.05] & 0.498 & 0.570 \\
Cross: HNSW(V)+BM25(D) & HNSW + BM25-FTS & W70 & 0.649 & 2.034 & [1.99, 2.08] & 0.502 & 0.580 \\
D Core+R & BM25-FTS & Add & 0.647 & 2.013 & [1.97, 2.05] & 0.500 & 0.570 \\
V & HNSW & --- & 0.645 & 2.002 & [1.96, 2.04] & 0.498 & 0.567 \\
Cross: HNSW(V)+BM25(D) & HNSW + BM25-Okapi & W95 & 0.645 & 2.002 & [1.96, 2.04] & 0.493 & 0.569 \\
Cross: HNSW(V)+BM25(D) & HNSW + BM25-FTS & W95 & 0.644 & 2.007 & [1.97, 2.05] & 0.493 & 0.570 \\
Cross: HNSW(V)+BM25(D) & HNSW + BM25-Okapi & W90 & 0.643 & 2.007 & [1.97, 2.05] & 0.493 & 0.569 \\
D Core & BM25-Okapi & --- & 0.639 & 1.918 & [1.88, 1.96] & 0.495 & 0.538 \\
Cross: HNSW(V)+BM25(D) & HNSW + BM25-FTS & W90 & 0.638 & 2.011 & [1.97, 2.05] & 0.488 & 0.571 \\
Cross: HNSW(V)+BM25(D) & HNSW + BM25-Okapi & Add & 0.638 & 2.017 & [1.98, 2.06] & 0.488 & 0.573 \\
Cross: HNSW(V)+BM25(D) & HNSW + BM25-Okapi & W50 & 0.638 & 2.017 & [1.98, 2.06] & 0.488 & 0.573 \\
V & Exact & --- & 0.638 & 2.008 & [1.97, 2.05] & 0.488 & 0.568 \\
D Core+R & BM25-FTS & RRF & 0.637 & 1.969 & [1.93, 2.01] & 0.485 & 0.556 \\
D Core+R & Exact & RRF & 0.635 & 1.986 & [1.95, 2.03] & 0.468 & 0.559 \\
Cross: HNSW(V)+BM25(D) & HNSW + BM25-Okapi & CMZ & 0.634 & 2.010 & [1.97, 2.05] & 0.478 & 0.569 \\
D Core+F & BM25-FTS & RRF & 0.627 & 1.924 & [1.88, 1.97] & 0.495 & 0.542 \\
D Core+R & HNSW & RRF & 0.626 & 1.975 & [1.93, 2.02] & 0.458 & 0.554 \\
Cross: HNSW(V)+BM25(D) & HNSW + BM25-Okapi & RRF & 0.625 & 2.010 & [1.97, 2.05] & 0.468 & 0.566 \\
D All & BM25-FTS & Add & 0.625 & 1.948 & [1.91, 1.99] & 0.485 & 0.548 \\
D Core+F & HNSW & Add & 0.624 & 1.897 & [1.85, 1.94] & 0.488 & 0.532 \\
D Core+F & HNSW & RRF & 0.623 & 1.913 & [1.87, 1.96] & 0.473 & 0.536 \\
D Core+F & Exact & Add & 0.622 & 1.924 & [1.88, 1.97] & 0.483 & 0.543 \\
D All & BM25-Okapi & Wt & 0.621 & 1.900 & [1.86, 1.94] & 0.488 & 0.531 \\
D Core+F & BM25-FTS & Add & 0.614 & 1.944 & [1.90, 1.99] & 0.470 & 0.546 \\
D Core+F & Exact & RRF & 0.612 & 1.918 & [1.88, 1.96] & 0.448 & 0.537 \\
D Core+R & BM25-Okapi & Add & 0.611 & 1.951 & [1.91, 1.99] & 0.450 & 0.547 \\
D All & HNSW & Add & 0.608 & 1.906 & [1.86, 1.95] & 0.448 & 0.534 \\
D All & Exact & Add & 0.608 & 1.921 & [1.88, 1.96] & 0.453 & 0.539 \\
D Core+R & BM25-Okapi & RRF & 0.605 & 1.907 & [1.86, 1.95] & 0.445 & 0.529 \\
D Core+F & BM25-Okapi & RRF & 0.598 & 1.852 & [1.81, 1.90] & 0.445 & 0.512 \\
D All & BM25-Okapi & Add & 0.589 & 1.862 & [1.82, 1.91] & 0.458 & 0.514 \\
D Core+F & BM25-Okapi & Add & 0.585 & 1.845 & [1.80, 1.89] & 0.445 & 0.513 \\
D All & BM25-FTS & RRF & 0.584 & 1.913 & [1.87, 1.96] & 0.445 & 0.530 \\
D All & HNSW & RRF & 0.570 & 1.922 & [1.88, 1.96] & 0.393 & 0.530 \\
D All & BM25-Okapi & RRF & 0.568 & 1.853 & [1.81, 1.89] & 0.428 & 0.503 \\
D All & Exact & RRF & 0.559 & 1.926 & [1.88, 1.97] & 0.378 & 0.529 \\
\end{longtable}
\endgroup

\subsection{Results by Query Type}
\begin{table}[htbp]
\centering
\caption{Representative query-type contrasts.}
\label{tab:query-type-summary}
\small
\begin{tabular}{p{2.2cm}p{3.9cm}c p{3.9cm}c}
\toprule
Query type & Stronger setup & Mean grade & Comparison setup & Mean grade \\
\midrule
Exact term & Core+Files / HNSW / Wt & \qtExactDistillCFHnswWt{} & Full Text / HNSW & \qtExactVerbatimHnsw{} \\
Conceptual & Full Text / HNSW & \qtConceptualVerbatimHnsw{} & Core+Files / HNSW / RRF & \qtConceptualDistillCFHnswRRF{} \\
\bottomrule
\end{tabular}
\end{table}
% Table 2: Results by Query Type
\begingroup
\footnotesize
\setlength{\tabcolsep}{3pt}
\begin{longtable}{lllccc}
\caption{Retrieval quality disaggregated by query type ($\pm$ 95\% CI). Row order follows Table~\ref{tab:main-results} (MRR descending). Query types: conceptual (abstract topics), phrase (multi-word patterns), exact term (specific identifiers).}
\label{tab:query-type} \\
\toprule
Mode & Mechanism & Fusion & Conceptual & Phrase & Exact Term \\
\midrule
\endfirsthead
\multicolumn{6}{l}{\small\itshape Table~\ref{tab:query-type} continued} \\
\toprule
Mode & Mechanism & Fusion & Conceptual & Phrase & Exact Term \\
\midrule
\endhead
\midrule
\multicolumn{6}{r}{\small\itshape Continued on next page} \\
\endfoot
\bottomrule
\endlastfoot
Cross: BM25(V)+HNSW(D) & BM25-FTS + HNSW & CombMNZ & $2.26 \pm 0.05$ & $2.12 \pm 0.07$ & $2.36 \pm 0.08$ \\
Cross: BM25(V)+HNSW(D) & BM25-FTS + HNSW & W70 & $2.29 \pm 0.05$ & $2.13 \pm 0.07$ & $2.40 \pm 0.08$ \\
Cross: BM25(V)+HNSW(D) & BM25-FTS + HNSW & W90 & $2.25 \pm 0.05$ & $2.10 \pm 0.08$ & $2.40 \pm 0.08$ \\
Cross: BM25(V)+HNSW(D) & BM25-Okapi + HNSW & W70 & $2.17 \pm 0.06$ & $2.10 \pm 0.08$ & $2.31 \pm 0.08$ \\
Cross: BM25(V)+HNSW(D) & BM25-FTS + HNSW & W95 & $2.25 \pm 0.05$ & $2.10 \pm 0.08$ & $2.38 \pm 0.08$ \\
Cross: BM25(V)+HNSW(D) & BM25-FTS + HNSW & Additive & $2.27 \pm 0.05$ & $2.13 \pm 0.07$ & $2.39 \pm 0.08$ \\
Cross: BM25(V)+HNSW(D) & BM25-FTS + HNSW & W50 & $2.27 \pm 0.05$ & $2.13 \pm 0.07$ & $2.39 \pm 0.08$ \\
Cross: 3-Signal & BM25-FTS×2 + HNSW & W90 & $2.25 \pm 0.05$ & $2.10 \pm 0.08$ & $2.38 \pm 0.08$ \\
Cross: BM25(V)+HNSW(D) & BM25-FTS + HNSW & RRF & $2.26 \pm 0.05$ & $2.12 \pm 0.07$ & $2.35 \pm 0.08$ \\
Cross: BM25(V)+HNSW(D) & BM25-Okapi + HNSW & W80 & $2.19 \pm 0.06$ & $2.07 \pm 0.08$ & $2.33 \pm 0.08$ \\
Cross: BM25(V)+HNSW(D) & BM25-FTS + HNSW & Max & $2.23 \pm 0.06$ & $2.10 \pm 0.08$ & $2.33 \pm 0.08$ \\
Cross: 3-Signal & BM25-FTS×2 + HNSW & W95 & $2.25 \pm 0.05$ & $2.10 \pm 0.08$ & $2.38 \pm 0.08$ \\
Full Text & BM25-FTS & --- & $2.25 \pm 0.05$ & $2.08 \pm 0.08$ & $2.38 \pm 0.08$ \\
Cross: 3-Signal & BM25-Okapi×2 + HNSW & W70 & $2.16 \pm 0.06$ & $2.03 \pm 0.08$ & $2.31 \pm 0.09$ \\
Cross: BM25(V)+HNSW(D) & BM25-Okapi + HNSW & CombMNZ & $2.21 \pm 0.05$ & $2.11 \pm 0.07$ & $2.27 \pm 0.09$ \\
Cross: BM25(V)+HNSW(D) & BM25-FTS + HNSW & W80 & $2.26 \pm 0.05$ & $2.11 \pm 0.08$ & $2.40 \pm 0.08$ \\
Cross: 3-Signal & BM25-Okapi×2 + HNSW & W80 & $2.16 \pm 0.06$ & $2.02 \pm 0.08$ & $2.33 \pm 0.08$ \\
Cross: 3-Signal & BM25-FTS×2 + HNSW & Max & $2.23 \pm 0.06$ & $2.05 \pm 0.08$ & $2.31 \pm 0.08$ \\
Cross: 3-Signal & BM25-FTS×2 + HNSW & W70 & $2.26 \pm 0.05$ & $2.09 \pm 0.08$ & $2.35 \pm 0.08$ \\
Cross: 3-Signal & BM25-FTS×2 + HNSW & W80 & $2.26 \pm 0.05$ & $2.11 \pm 0.08$ & $2.37 \pm 0.08$ \\
Cross: BM25(V)+HNSW(D) & BM25-Okapi + HNSW & RRF & $2.19 \pm 0.05$ & $2.11 \pm 0.08$ & $2.29 \pm 0.09$ \\
Cross: BM25(V)+HNSW(D) & BM25-Okapi + HNSW & Additive & $2.20 \pm 0.05$ & $2.10 \pm 0.08$ & $2.30 \pm 0.09$ \\
Cross: BM25(V)+HNSW(D) & BM25-Okapi + HNSW & W50 & $2.20 \pm 0.05$ & $2.10 \pm 0.08$ & $2.30 \pm 0.09$ \\
Cross: 3-Signal & BM25-FTS×2 + HNSW & W50 & $2.28 \pm 0.05$ & $2.08 \pm 0.08$ & $2.34 \pm 0.08$ \\
Cross: 3-Signal & BM25-Okapi×2 + HNSW & W90 & $2.16 \pm 0.06$ & $2.04 \pm 0.08$ & $2.31 \pm 0.08$ \\
Cross: 3-Signal & BM25-Okapi×2 + HNSW & Max & $2.14 \pm 0.06$ & $2.00 \pm 0.08$ & $2.23 \pm 0.09$ \\
Cross: BM25(V)+HNSW(D) & BM25-Okapi + HNSW & W90 & $2.18 \pm 0.06$ & $2.05 \pm 0.08$ & $2.32 \pm 0.08$ \\
Cross: BM25(V)+HNSW(D) & BM25-Okapi + HNSW & W95 & $2.17 \pm 0.06$ & $2.04 \pm 0.08$ & $2.31 \pm 0.08$ \\
Cross: BM25(V)+HNSW(D) & BM25-Okapi + HNSW & Max & $2.16 \pm 0.06$ & $2.08 \pm 0.08$ & $2.27 \pm 0.09$ \\
Cross: 3-Signal & BM25-Okapi×2 + HNSW & W50 & $2.14 \pm 0.06$ & $2.05 \pm 0.08$ & $2.27 \pm 0.09$ \\
Cross: 3-Signal & BM25-Okapi×2 + HNSW & W95 & $2.15 \pm 0.06$ & $2.05 \pm 0.08$ & $2.31 \pm 0.08$ \\
Full Text & BM25-Okapi & --- & $2.16 \pm 0.06$ & $2.03 \pm 0.08$ & $2.30 \pm 0.09$ \\
Distill Core+Rooms & Exact & Weighted & $2.13 \pm 0.06$ & $1.95 \pm 0.08$ & $2.14 \pm 0.09$ \\
Cross: BM25(V)+HNSW(D) rev & HNSW + BM25-FTS & W70 & $2.21 \pm 0.05$ & $2.05 \pm 0.07$ & $2.23 \pm 0.09$ \\
Cross: 3-Signal & BM25-FTS×2 + HNSW & Additive & $2.23 \pm 0.05$ & $2.05 \pm 0.08$ & $2.29 \pm 0.09$ \\
Cross: 3-Signal & BM25-FTS×2 + HNSW & CombMNZ & $2.20 \pm 0.05$ & $2.03 \pm 0.08$ & $2.25 \pm 0.09$ \\
Cross: BM25(V)+HNSW(D) rev & HNSW + BM25-FTS & W95 & $2.12 \pm 0.06$ & $2.00 \pm 0.08$ & $2.15 \pm 0.09$ \\
Cross: BM25(V)+HNSW(D) rev & HNSW + BM25-FTS & W80 & $2.17 \pm 0.06$ & $2.03 \pm 0.08$ & $2.19 \pm 0.09$ \\
Cross: BM25(V)+HNSW(D) rev & HNSW + BM25-FTS & W90 & $2.13 \pm 0.06$ & $2.01 \pm 0.08$ & $2.17 \pm 0.09$ \\
Distill Core & Exact & --- & $2.11 \pm 0.06$ & $1.97 \pm 0.08$ & $2.15 \pm 0.09$ \\
Cross: 3-Signal & BM25-FTS×2 + HNSW & RRF & $2.19 \pm 0.05$ & $2.00 \pm 0.08$ & $2.26 \pm 0.09$ \\
Distill Core+Rooms & HNSW & Weighted & $2.12 \pm 0.06$ & $1.95 \pm 0.08$ & $2.14 \pm 0.09$ \\
Cross: 3-Signal & BM25-Okapi×2 + HNSW & Additive & $2.16 \pm 0.06$ & $2.02 \pm 0.08$ & $2.23 \pm 0.09$ \\
Distill Core+Files & Exact & Weighted & $2.09 \pm 0.06$ & $1.93 \pm 0.08$ & $2.11 \pm 0.09$ \\
Distill Core & HNSW & --- & $2.11 \pm 0.06$ & $1.99 \pm 0.08$ & $2.14 \pm 0.09$ \\
Cross: 3-Signal & BM25-Okapi×2 + HNSW & CombMNZ & $2.12 \pm 0.06$ & $1.97 \pm 0.08$ & $2.21 \pm 0.09$ \\
Distill Core+Files & HNSW & Weighted & $2.07 \pm 0.06$ & $1.94 \pm 0.08$ & $2.08 \pm 0.09$ \\
Cross: 3-Signal & BM25-Okapi×2 + HNSW & RRF & $2.12 \pm 0.06$ & $1.98 \pm 0.08$ & $2.20 \pm 0.09$ \\
Cross: HNSW(V)+BM25(D) rev & BM25-FTS + HNSW & W70 & $2.13 \pm 0.06$ & $1.94 \pm 0.08$ & $2.07 \pm 0.09$ \\
Distill All Facets & Exact & Weighted & $2.05 \pm 0.06$ & $1.92 \pm 0.08$ & $2.15 \pm 0.09$ \\
Distill All Facets & HNSW & Weighted & $2.04 \pm 0.06$ & $1.94 \pm 0.08$ & $2.12 \pm 0.09$ \\
Distill Core+Files & BM25-FTS & Weighted & $2.08 \pm 0.06$ & $1.86 \pm 0.08$ & $2.15 \pm 0.09$ \\
Cross: HNSW(V)+BM25(D) rev & BM25-FTS + HNSW & W50 & $2.13 \pm 0.06$ & $1.99 \pm 0.08$ & $2.04 \pm 0.10$ \\
Cross: HNSW(V)+BM25(D) rev & BM25-FTS + HNSW & W80 & $2.10 \pm 0.06$ & $1.90 \pm 0.08$ & $2.08 \pm 0.09$ \\
Cross: HNSW(V)+BM25(D) rev & BM25-FTS + HNSW & W95 & $2.09 \pm 0.06$ & $1.87 \pm 0.08$ & $2.09 \pm 0.09$ \\
Cross: HNSW(V)+BM25(D) & HNSW + BM25-FTS & RRF & $2.10 \pm 0.06$ & $1.97 \pm 0.08$ & $2.10 \pm 0.09$ \\
Distill Core+Rooms & BM25-FTS & Weighted & $2.11 \pm 0.06$ & $1.84 \pm 0.08$ & $2.17 \pm 0.09$ \\
Distill Core & BM25-FTS & --- & $2.08 \pm 0.06$ & $1.86 \pm 0.08$ & $2.15 \pm 0.10$ \\
Cross: HNSW(V)+BM25(D) & HNSW + BM25-FTS & Additive & $2.12 \pm 0.06$ & $1.99 \pm 0.08$ & $2.03 \pm 0.10$ \\
Cross: HNSW(V)+BM25(D) & HNSW + BM25-FTS & W50 & $2.12 \pm 0.06$ & $1.99 \pm 0.08$ & $2.03 \pm 0.10$ \\
Cross: HNSW(V)+BM25(D) rev & BM25-FTS + HNSW & W90 & $2.10 \pm 0.06$ & $1.88 \pm 0.08$ & $2.08 \pm 0.09$ \\
Cross: HNSW(V)+BM25(D) & HNSW + BM25-FTS & CombMNZ & $2.12 \pm 0.06$ & $2.01 \pm 0.08$ & $2.05 \pm 0.10$ \\
Distill Core+Files & BM25-Okapi & Weighted & $1.97 \pm 0.06$ & $1.71 \pm 0.08$ & $2.14 \pm 0.09$ \\
Cross: HNSW(V)+BM25(D) & HNSW + BM25-Okapi & W70 & $2.08 \pm 0.06$ & $1.94 \pm 0.07$ & $1.97 \pm 0.10$ \\
Distill Core+Rooms & Exact & Additive & $2.07 \pm 0.06$ & $1.84 \pm 0.08$ & $2.07 \pm 0.09$ \\
Cross: HNSW(V)+BM25(D) & HNSW + BM25-FTS & W80 & $2.08 \pm 0.06$ & $1.94 \pm 0.07$ & $1.97 \pm 0.10$ \\
Distill All Facets & BM25-FTS & Weighted & $2.05 \pm 0.06$ & $1.80 \pm 0.08$ & $2.14 \pm 0.09$ \\
Distill Core+Rooms & HNSW & Additive & $2.06 \pm 0.06$ & $1.85 \pm 0.08$ & $2.04 \pm 0.10$ \\
Cross: HNSW(V)+BM25(D) & HNSW + BM25-Okapi & W80 & $2.06 \pm 0.06$ & $1.93 \pm 0.07$ & $1.96 \pm 0.10$ \\
Distill Core+Rooms & BM25-Okapi & Weighted & $2.02 \pm 0.06$ & $1.76 \pm 0.08$ & $2.12 \pm 0.09$ \\
Cross: HNSW(V)+BM25(D) & HNSW + BM25-FTS & Max & $2.11 \pm 0.06$ & $1.98 \pm 0.08$ & $2.00 \pm 0.10$ \\
Cross: HNSW(V)+BM25(D) & HNSW + BM25-Okapi & Max & $2.07 \pm 0.06$ & $1.92 \pm 0.07$ & $2.01 \pm 0.10$ \\
Cross: HNSW(V)+BM25(D) & HNSW + BM25-FTS & W70 & $2.09 \pm 0.06$ & $1.97 \pm 0.07$ & $1.99 \pm 0.10$ \\
Distill Core+Rooms & BM25-FTS & Additive & $2.08 \pm 0.06$ & $1.81 \pm 0.08$ & $2.16 \pm 0.09$ \\
Full Text & HNSW & --- & $2.08 \pm 0.06$ & $1.92 \pm 0.07$ & $1.93 \pm 0.10$ \\
Cross: HNSW(V)+BM25(D) & HNSW + BM25-Okapi & W95 & $2.08 \pm 0.06$ & $1.93 \pm 0.07$ & $1.92 \pm 0.10$ \\
Cross: HNSW(V)+BM25(D) & HNSW + BM25-FTS & W95 & $2.08 \pm 0.06$ & $1.94 \pm 0.07$ & $1.93 \pm 0.10$ \\
Cross: HNSW(V)+BM25(D) & HNSW + BM25-Okapi & W90 & $2.08 \pm 0.06$ & $1.93 \pm 0.07$ & $1.94 \pm 0.10$ \\
Distill Core & BM25-Okapi & --- & $1.96 \pm 0.06$ & $1.73 \pm 0.08$ & $2.10 \pm 0.09$ \\
Cross: HNSW(V)+BM25(D) & HNSW + BM25-FTS & W90 & $2.08 \pm 0.06$ & $1.94 \pm 0.07$ & $1.95 \pm 0.10$ \\
Cross: HNSW(V)+BM25(D) & HNSW + BM25-Okapi & Additive & $2.09 \pm 0.06$ & $1.92 \pm 0.07$ & $1.99 \pm 0.10$ \\
Cross: HNSW(V)+BM25(D) & HNSW + BM25-Okapi & W50 & $2.09 \pm 0.06$ & $1.92 \pm 0.07$ & $1.99 \pm 0.10$ \\
Full Text & Exact & --- & $2.08 \pm 0.06$ & $1.93 \pm 0.07$ & $1.94 \pm 0.10$ \\
Distill Core+Rooms & BM25-FTS & RRF & $2.05 \pm 0.06$ & $1.76 \pm 0.08$ & $2.10 \pm 0.10$ \\
Distill Core+Rooms & Exact & RRF & $2.06 \pm 0.06$ & $1.84 \pm 0.08$ & $2.02 \pm 0.09$ \\
Cross: HNSW(V)+BM25(D) & HNSW + BM25-Okapi & CombMNZ & $2.08 \pm 0.06$ & $1.89 \pm 0.08$ & $2.02 \pm 0.10$ \\
Distill Core+Files & BM25-FTS & RRF & $1.95 \pm 0.06$ & $1.73 \pm 0.08$ & $2.17 \pm 0.09$ \\
Distill Core+Rooms & HNSW & RRF & $2.05 \pm 0.06$ & $1.83 \pm 0.08$ & $2.00 \pm 0.09$ \\
Cross: HNSW(V)+BM25(D) & HNSW + BM25-Okapi & RRF & $2.06 \pm 0.06$ & $1.88 \pm 0.08$ & $2.08 \pm 0.09$ \\
Distill All Facets & BM25-FTS & Additive & $2.01 \pm 0.06$ & $1.71 \pm 0.08$ & $2.16 \pm 0.09$ \\
Distill Core+Files & HNSW & Additive & $1.93 \pm 0.06$ & $1.76 \pm 0.08$ & $2.02 \pm 0.09$ \\
Distill Core+Files & HNSW & RRF & $1.96 \pm 0.06$ & $1.80 \pm 0.08$ & $1.97 \pm 0.09$ \\
Distill Core+Files & Exact & Additive & $1.96 \pm 0.06$ & $1.78 \pm 0.08$ & $2.06 \pm 0.09$ \\
Distill All Facets & BM25-Okapi & Weighted & $1.96 \pm 0.06$ & $1.67 \pm 0.08$ & $2.09 \pm 0.09$ \\
Distill Core+Files & BM25-FTS & Additive & $1.97 \pm 0.06$ & $1.75 \pm 0.08$ & $2.19 \pm 0.09$ \\
Distill Core+Files & Exact & RRF & $1.97 \pm 0.06$ & $1.77 \pm 0.08$ & $2.01 \pm 0.09$ \\
Distill Core+Rooms & BM25-Okapi & Additive & $2.03 \pm 0.06$ & $1.74 \pm 0.08$ & $2.09 \pm 0.09$ \\
Distill All Facets & HNSW & Additive & $1.93 \pm 0.06$ & $1.78 \pm 0.08$ & $2.06 \pm 0.09$ \\
Distill All Facets & Exact & Additive & $1.95 \pm 0.06$ & $1.79 \pm 0.08$ & $2.06 \pm 0.09$ \\
Distill Core+Rooms & BM25-Okapi & RRF & $2.00 \pm 0.06$ & $1.68 \pm 0.08$ & $2.02 \pm 0.09$ \\
Distill Core+Files & BM25-Okapi & RRF & $1.88 \pm 0.06$ & $1.63 \pm 0.08$ & $2.12 \pm 0.09$ \\
Distill All Facets & BM25-Okapi & Additive & $1.93 \pm 0.06$ & $1.60 \pm 0.08$ & $2.09 \pm 0.09$ \\
Distill Core+Files & BM25-Okapi & Additive & $1.88 \pm 0.06$ & $1.59 \pm 0.08$ & $2.17 \pm 0.09$ \\
Distill All Facets & BM25-FTS & RRF & $1.98 \pm 0.06$ & $1.71 \pm 0.08$ & $2.07 \pm 0.09$ \\
Distill All Facets & HNSW & RRF & $1.96 \pm 0.06$ & $1.81 \pm 0.08$ & $2.00 \pm 0.09$ \\
Distill All Facets & BM25-Okapi & RRF & $1.93 \pm 0.06$ & $1.63 \pm 0.08$ & $2.01 \pm 0.09$ \\
Distill All Facets & Exact & RRF & $1.98 \pm 0.06$ & $1.80 \pm 0.08$ & $1.99 \pm 0.09$ \\
\end{longtable}
\endgroup

\subsection{Inter-Rater Agreement Details}
% Table 4: Inter-Rater Agreement (auto-generated)
\begin{table}[htbp]
\centering
\caption{Inter-rater agreement across 5 LLM graders. Despite low absolute $\kappa$, the mechanism-dependent preservation finding is directionally consistent across all graders. Interpretation follows Landis \& Koch (1977): $\kappa < 0.20$ = Slight, $0.20$--$0.40$ = Fair.}
\label{tab:agreement}
\footnotesize
\setlength{\tabcolsep}{4pt}
\begin{tabular}{p{0.62\columnwidth}r}
\toprule
Measure & Value \\
\midrule
Fleiss' $\kappa$ (all 5 graders) & 0.175 (Slight) \\
Mean pairwise Cohen's $\kappa$ & 0.192 (Slight) \\
Best pair (Phi-3.5-Mini $\leftrightarrow$ Qwen3-8B) & 0.260 (Fair) \\
Worst pair (Mistral-7B $\leftrightarrow$ Phi-3.5-Mini) & 0.131 (Slight) \\
Items rated by all 5 graders & 3,672 \\
\bottomrule
\end{tabular}
\end{table}

\subsection{Statistical Test Details}
% Table 3: Per-Mechanism Statistical Comparisons (40 total, auto-generated)
\begingroup
\footnotesize
\setlength{\tabcolsep}{4pt}
\begin{longtable}{lllrrrl}
\caption{Mechanism-dependent quality preservation: per-mechanism pairwise comparisons (40 total). Baseline: Full Text (trivial fusion). Vector search (HNSW, Exact) shows non-significant degradation; keyword search (BM25) shows consistent significant degradation. Bonferroni-corrected $\alpha = 0.00125$. $^\dagger$ = best distilled MRR for this mechanism. *** = significant after Bonferroni ($p < 0.00125$). ** = significant at nominal $\alpha = 0.05$ but not after Bonferroni correction. V = Full Text, D = Distill, F = Files, R = Rooms. Add = Additive, Wt = Weighted. Effect: N = negligible, S = small, M = medium.}
\label{tab:stat-tests} \\
\toprule
Mechanism & Mode & Fusion & $\Delta$ & $t$ & $p$ & $d$ \\
\midrule
\endfirsthead
\multicolumn{7}{l}{\small\itshape Table~\ref{tab:stat-tests} continued} \\
\toprule
Mechanism & Mode & Fusion & $\Delta$ & $t$ & $p$ & $d$ \\
\midrule
\endhead
\midrule
\multicolumn{7}{r}{\small\itshape Continued on next page} \\
\endfoot
\bottomrule
\endlastfoot
\multicolumn{7}{l}{\textbf{HNSW}} \\
& D Core$^\dagger$ & --- & $-$0.052 & $-$1.93 & 0.055 & $-$0.14N \\
& D Core+F & Add & +0.105 & 3.07 & 0.0025** & 0.22S \\
& D Core+F & RRF & +0.089 & 2.73 & 0.0068** & 0.19N \\
& D Core+F$^\dagger$ & Wt & $-$0.028 & $-$1.00 & 0.319 & $-$0.07N \\
& D Core+R & Add & +0.014 & 0.52 & 0.601 & 0.04N \\
& D Core+R & RRF & +0.030 & 1.12 & 0.263 & 0.08N \\
& D Core+R$^\dagger$ & Wt & $-$0.065 & $-$2.41 & 0.0168** & $-$0.17N \\
& D All & Add & +0.096 & 2.99 & 0.0031** & 0.21S \\
& D All & RRF & +0.080 & 2.80 & 0.0057** & 0.20N \\
& D All$^\dagger$ & Wt & $-$0.020 & $-$0.73 & 0.466 & $-$0.05N \\
\midrule
\multicolumn{7}{l}{\textbf{Exact}} \\
& D Core$^\dagger$ & --- & $-$0.061 & $-$2.25 & 0.0253** & $-$0.16N \\
& D Core+F & Add & +0.084 & 2.52 & 0.0126** & 0.18N \\
& D Core+F & RRF & +0.090 & 2.76 & 0.0063** & 0.19N \\
& D Core+F$^\dagger$ & Wt & $-$0.040 & $-$1.44 & 0.152 & $-$0.10N \\
& D Core+R & Add & +0.012 & 0.44 & 0.662 & 0.03N \\
& D Core+R & RRF & +0.023 & 0.86 & 0.392 & 0.06N \\
& D Core+R$^\dagger$ & Wt & $-$0.067 & $-$2.57 & 0.0109** & $-$0.18N \\
& D All & Add & +0.088 & 2.76 & 0.0064** & 0.19N \\
& D All & RRF & +0.083 & 2.89 & 0.0043** & 0.20S \\
& D All$^\dagger$ & Wt & $-$0.023 & $-$0.83 & 0.409 & $-$0.06N \\
\midrule
\multicolumn{7}{l}{\textbf{BM25-Okapi}} \\
& D Core$^\dagger$ & --- & +0.238 & 8.67 & $2\mathrm{e}-15$*** & 0.61M \\
& D Core+F & Add & +0.306 & 9.26 & $3\mathrm{e}-17$*** & 0.66M \\
& D Core+F & RRF & +0.299 & 9.59 & $4\mathrm{e}-18$*** & 0.68M \\
& D Core+F$^\dagger$ & Wt & +0.225 & 7.82 & $3\mathrm{e}-13$*** & 0.55M \\
& D Core+R & Add & +0.200 & 6.44 & $9\mathrm{e}-10$*** & 0.46S \\
& D Core+R & RRF & +0.245 & 8.22 & $3\mathrm{e}-14$*** & 0.58M \\
& D Core+R$^\dagger$ & Wt & +0.191 & 6.86 & $8\mathrm{e}-11$*** & 0.49S \\
& D All & Add & +0.287 & 8.90 & $3\mathrm{e}-16$*** & 0.63M \\
& D All & RRF & +0.297 & 9.28 & $3\mathrm{e}-17$*** & 0.65M \\
& D All$^\dagger$ & Wt & +0.250 & 8.51 & $4\mathrm{e}-15$*** & 0.60M \\
\midrule
\multicolumn{7}{l}{\textbf{BM25-FTS}} \\
& D Core$^\dagger$ & --- & +0.202 & 8.08 & $6\mathrm{e}-14$*** & 0.57M \\
& D Core+F & Add & +0.283 & 9.29 & $3\mathrm{e}-17$*** & 0.66M \\
& D Core+F & RRF & +0.303 & 10.16 & $8\mathrm{e}-20$*** & 0.72M \\
& D Core+F$^\dagger$ & Wt & +0.203 & 7.85 & $2\mathrm{e}-13$*** & 0.56M \\
& D Core+R & Add & +0.215 & 7.74 & $5\mathrm{e}-13$*** & 0.55M \\
& D Core+R & RRF & +0.258 & 8.75 & $9\mathrm{e}-16$*** & 0.62M \\
& D Core+R$^\dagger$ & Wt & +0.187 & 7.25 & $9\mathrm{e}-12$*** & 0.51M \\
& D All & Add & +0.279 & 9.45 & $10\mathrm{e}-18$*** & 0.67M \\
& D All & RRF & +0.314 & 10.70 & $2\mathrm{e}-21$*** & 0.76M \\
& D All$^\dagger$ & Wt & +0.237 & 8.88 & $4\mathrm{e}-16$*** & 0.63M \\
\end{longtable}
\endgroup

\section{Distillation Prompt}
\label{app:distillation-prompt}

The following prompt is used for batch distillation (Haiku), which produced the corpus evaluated in this paper.
\texttt{\{project\_id\}}, \texttt{\{ply\_start\}}, \texttt{\{ply\_end\}}, and \texttt{\{messages\_text\}} are interpolated per exchange.
Messages are truncated to \messageTruncChars{} characters.

\begin{quote}
\small
\begin{verbatim}
Distill this conversation exchange into JSON:

- "exchange_core": 1-2 sentences. What was accomplished or decided?
  Use the specific terms from the exchange. Do not invent details
  not present in the text. If the exchange is mostly empty, say so
  briefly.
- "specific_context": One concrete detail from the text: a number,
  error message, parameter name, or file path. Copy it exactly from
  the text. Do not use the project path.
- "room_assignments": 1-3 rooms. Each room is a topic this exchange
  belongs to. {"room_type": "<file|concept|workflow>",
  "room_key": "<identifier>", "room_label": "<short label>",
  "relevance": <0.0-1.0>}. A room should be specific enough to
  group related exchanges (e.g. "retry_timeout" not "errors").

Do NOT include "files_touched".

Project: {project_id}

Exchange (messages {ply_start}-{ply_end}):
{messages_text}

Respond with ONLY valid JSON.
\end{verbatim}
\end{quote}

Per-turn distillation (Haiku) uses a simpler prompt that extracts \texttt{tags} (2--4 keywords) instead of \texttt{room\_assignments}.
The per-turn prompt runs as a Claude Code stop hook with a 60-second timeout.
Since the batch distiller re-processes the full corpus, the evaluated palace objects reflect the batch prompt above.

\section{Grading Prompt}
\label{app:grading-prompt}

The following prompt template is used for LLM-based relevance grading.
\texttt{\{query\}} and \texttt{\{snippet\}} are interpolated per item.

\begin{quote}
\small
\begin{verbatim}
You are a strict relevance assessor for a conversational
search system.

QUERY: "{query}"

SEARCH RESULT:
---
{snippet}
---

Grade this result's relevance to the query on a 0-3 scale.

SCALE:
0 = Irrelevant: The result has nothing to do with the query.
    Different topic entirely.

1 = Related: The result is on a related topic but does NOT
    answer the query. It may mention similar concepts or
    share terminology, but a user would not find what they
    were looking for.

2 = Highly Relevant: The result contains an answer or useful
    information for the query, but the answer may be unclear,
    incomplete, or buried among other content.

3 = Perfectly Relevant: The result is dedicated to the query
    topic and directly provides the information sought.

GRADING RULES:
- The key test for grade 2 vs grade 1: does the result
  contain a usable answer? If yes, grade 2+. If it is merely
  on a related topic, grade 1.
- Only assign grade 3 if the result specifically and clearly
  addresses the query.
- When uncertain between two grades, assign the lower one.
- A result about a related but different tool, version, or
  concept is grade 1, not grade 2.

Respond with a JSON object. Write your reasoning first,
then the grade.
{"reason": "<10-20 word justification>", "grade": <0|1|2|3>}
\end{verbatim}
\end{quote}

\section{Best Cases: Distilled Modes Outperform Verbatim}
\label{app:best-cases}

The following queries show the largest advantage for distilled search over a matched Full Text baseline, measured as the difference in mean grade between the best distilled configuration and Full Text with the same retrieval mechanism.

\paragraph{Query A (\bestCaseOneQueryType):} Distilled advantage: \bestCaseOneDiff{} grade points (\bestCaseOneModeLabel{} / \bestCaseOneMechanismLabel{} / \bestCaseOneFusionLabel{} mean \bestCaseOneDistilledMean{}, Full Text mean \bestCaseOneVerbatimMean{}).
The gain comes from stripping away long exchange padding while preserving the specific term the user is trying to recall.

\paragraph{Query B (\bestCaseTwoQueryType):} Distilled advantage: \bestCaseTwoDiff{} grade points (\bestCaseTwoModeLabel{} / \bestCaseTwoMechanismLabel{} / \bestCaseTwoFusionLabel{} mean \bestCaseTwoDistilledMean{}, Full Text mean \bestCaseTwoVerbatimMean{}).
This is the same pattern in a more extreme form: the verbatim hit exists, but it is weak because the target term is buried inside surrounding material.

\paragraph{Query C (\bestCaseThreeQueryType):} Distilled advantage: \bestCaseThreeDiff{} grade points (\bestCaseThreeModeLabel{} / \bestCaseThreeMechanismLabel{} / \bestCaseThreeFusionLabel{} mean \bestCaseThreeDistilledMean{}, Full Text mean \bestCaseThreeVerbatimMean{}).
The compressed representation keeps the decisive phrase near the top of the object, which makes retrieval less sensitive to the noise of the original exchange.

\paragraph{Query D (\bestCaseFourQueryType):} Distilled advantage: \bestCaseFourDiff{} grade points (\bestCaseFourModeLabel{} / \bestCaseFourMechanismLabel{} / \bestCaseFourFusionLabel{} mean \bestCaseFourDistilledMean{}, Full Text mean \bestCaseFourVerbatimMean{}).
The winning configuration here adds structure without losing the local term match. That combination is exactly what the personalized memory layer is trying to buy.

\paragraph{Query E (\bestCaseFiveQueryType):} Distilled advantage: \bestCaseFiveDiff{} grade points (\bestCaseFiveModeLabel{} / \bestCaseFiveMechanismLabel{} / \bestCaseFiveFusionLabel{} mean \bestCaseFiveDistilledMean{}, Full Text mean \bestCaseFiveVerbatimMean{}).
This case is smaller, but it points the same way. Distillation helps when the remembered clue is real but thinly distributed in the verbatim exchange history.

Distilled search outperforms verbatim when specific terms or concepts are buried in long exchanges.
Distillation surfaces the essential content, while structured metadata (files + rooms) provides precise retrieval signals that verbatim text lacks.
The advantages are largest for exact-term queries where the target phrase appears tangentially in verbatim text but centrally in the distilled exchange\_core.

\section{Failure Cases: All Modes Struggle}
\label{app:failure-cases}

The following queries have the lowest mean grade across all evaluated configurations.

\paragraph{Query F (\failureCaseOneQueryType):} Mean grade across all configurations: \failureCaseOneMean{}.
The user is asking for something the corpus does not contain in a stable, directly retrievable form.

\paragraph{Query G (\failureCaseTwoQueryType):} Mean grade across all configurations: \failureCaseTwoMean{}.
There are topical neighbors, but not a strong enough exchange to anchor the query cleanly.

\paragraph{Query H (\failureCaseThreeQueryType):} Mean grade across all configurations: \failureCaseThreeMean{}.
This is a low-signal recall target. The system finds adjacent material, not the thing itself.

\paragraph{Query I (\failureCaseFourQueryType):} Mean grade across all configurations: \failureCaseFourMean{}.
The underlying idea exists in the history, but not in the framing the query asks for.

\paragraph{Query J (\failureCaseFiveQueryType):} Mean grade across all configurations: \failureCaseFiveMean{}.
A vague or weakly grounded phrase produces broad topical matches rather than a clear hit.

These failure cases share a common pattern: the queries target concepts that either (a) were not substantively discussed in the corpus, (b) were discussed with different terminology, or (c) are too generic to distinguish the relevant exchange from topically related noise.
Distillation cannot improve retrieval for content that does not exist in the corpus or that lacks distinctive vocabulary.

\end{document}